\documentclass{elsarticle}

\pdfoutput=1
\usepackage{graphicx} 
\usepackage{geometry} 
\usepackage{amsmath} 
\usepackage{amssymb} 
\usepackage[utf8x]{inputenc} 
\usepackage[T1]{fontenc}
\usepackage{hyperref}
\usepackage{booktabs}  
\usepackage{multirow}
\usepackage{subfig}
\usepackage[linesnumbered]{algorithm2e}
\usepackage{textcomp} 

\usepackage[normalem]{ulem}
\useunder{\uline}{\ul}{}

\makeatletter
\def\ps@pprintTitle{%
 \let\@oddhead\@empty
 \let\@evenhead\@empty
 \def\@oddfoot{Accepted Manuscript. DOI: 10.1016/j.cor.2017.04.012\hfill}%
 \let\@evenfoot\@oddfoot}
\makeatother

\usepackage{color}
\definecolor{darkorange}{rgb}{.71,0.21,0.01}
\definecolor{darkgreen}{rgb}{.12,.54,.11}
\definecolor{darkred}{rgb}{.8,0.0,0.0}
\hypersetup{
  breaklinks=true,  
  colorlinks=true,
  urlcolor=blue,
  linkcolor=darkorange,
  citecolor=darkgreen,
}
\newcommand{\R}{}

\begin{document}

\title{An improved Ant Colony System for the Sequential Ordering Problem}
\author[uos]{Rafa{\l} Skinderowicz}
\ead{rafal.skinderowicz@us.edu.pl}

\address[uos]{University of Silesia, Intitute of Computer
Science,\\B\k{e}dzi\'nska 39, 41-205 Sosnowiec, Poland\\
\vspace{1em}
       {\rm
       \textcopyright 2017. This manuscript version is made available under the CC-BY-NC-ND 4.0 license http://creativecommons.org/licenses/by-nc-nd/4.0/
       }
}

\begin{abstract}
It is not rare that the performance of one metaheuristic algorithm can be
improved by incorporating ideas taken from another.
In this article we present how Simulated Annealing (SA) can
be used to improve the efficiency of the Ant Colony System (ACS) and
Enhanced ACS when solving the Sequential Ordering Problem (SOP).
Moreover, we show how the very same ideas can be applied to improve the
convergence of a dedicated local search, i.e. the SOP-3-exchange algorithm.
A statistical analysis of the proposed algorithms both in terms of
finding suitable parameter values and the quality of the generated solutions is
presented based
on a series of computational experiments conducted on SOP instances from the
well-known TSPLIB and SOPLIB2006 repositories.
The proposed ACS-SA and EACS-SA algorithms often generate
solutions of better quality than the ACS and EACS, respectively.
Moreover, the EACS-SA algorithm combined with the proposed SOP-3-exchange-SA
    local search was able to find 10 new best solutions for the SOP
instances from the SOPLIB2006 repository, thus improving the state-of-the-art results
as known from the literature.
Overall, the best known or improved solutions
were found in 41 out of 48 cases.
\end{abstract}


\maketitle

\section{Introduction}
\label{sec:Introduction}

In recent years, a large number of metaheuristic optimization algorithms (MOAs) has been proposed, and some of these were created based on inspiration drawn from natural phenomena~\cite{glover2006}. Examples of these are the Ant Colony System algorithm that was inspired by the foraging behavior of certain species of ants and Simulated Annealing (SA) with some ideas taken from metallurgy~\cite{a6,czech2009}.
Metaheuristics are often applied to find solutions of an acceptable quality to difficult combinatorial optimization problems, particularly NP-complete ones.
A good example is the Sequential Ordering Problem (SOP)
 which consists in finding a minimum weight Hamiltonian path on a directed graph with weights that is subject to precedence constraints among the nodes. 
Although less time-consuming than the exact approaches, MOAs differ in their efficiency, which can sometimes be improved by combining ideas taken from other MOAs.

\subsection{Contributions}
\label{sec:Contributions}

The main aim of the paper is to show that Simulated Annealing could be used to improve the convergence speed of ACS and Enhanced ACS (EACS) algorithms. The proposed solution is easy to implement and does not increase the algorithms' asymptotic complexity.
Moreover, we developed a modified version of the \emph{SOP-3-exchange} local search (LS) heuristic as proposed by Gambardella et al.~\cite{a11} for the SOP. The modification, again, includes ideas taken from the SA to allow the algorithm to escape local optima. 
A thorough experimental evaluation on a number of SOP instances from well-known datasets confirms the efficiency of the proposed algorithms.
In fact, in several cases we obtained results of a better quality than those from state-of-the-art methods in the literature~\cite{a12,a13}.

The paper is organized as follows: Section~\ref{sec:related-work} focuses on recent ideas of improving the efficiency of the ACS (and related algorithms), some of which include the SA. We also briefly discuss recent work on solving the SOP. In Sec.~\ref{sec:improving} we describe our approach to improve the convergence speed of the ACS by using the SA.
Section~\ref{sec:local-search} presents a similar approach, but to improve the local minima escape ability of the SOP-3-exchange local search algorithm which is paired with the ACS and the EACS when solving the SOP.
Section~\ref{sec:experiments} presents the results of computational experiments conducted on two sets of SOP instances. The last section contains the conclusions and some ideas for further work.

\section{Related work}
\label{sec:Related_work}

\label{sec:related-work}

Multiple modifications to the ACO family of algorithms have been proposed in the literature. Most of them refer to pheromone update rules and parameter tuning.
Hassani et al.~\cite{a34} proposed a modified global pheromone update rule for the ACS in which not only the global best ant but also all ants with inferior solutions may update the pheromone with probability calculated according to the acceptance rule of the SA. The initial temperature was set arbitrarily to 100 and an exponential cooling schedule was applied. The limited computational experiments showed that in most cases the algorithm achieved  better results than the Ant System.
Bouhafs et al.~\cite{a33} proposed a two-phase approach based on the SA and ACS to solve the Capacitated Location-Routing Problem. The SA was used to find facility locations while the ACS was used to solve the corresponding location routing problem. In most cases The algorithm was able to improve the best-known solutions to the problem.

In most of the ACO and SA combinations the latter plays the role of a local search used to improve the quality of the solutions generated by the ants. 
Behnamian et al.~\cite{a42} proposed a hybrid of the ACO, SA and Variable Neighborhood Search algorithms for solving parallel-machine scheduling problems. The SA was used to guide the dedicated LS.
A successful combination of the ACS and the SA was proposed by Ayob and Jaradat~\cite{a35} for solving 
course timetabling problems. The SA was used along with the Tabu Search to improve solutions generated by the ACS. The results for the proposed algorithm were of better quality than those of the ACS alone or of the MAX-MIN Ant System.
The SA was again used as the LS for the ACS by Wassila and Boukra~\cite{a36}. The approach was slightly faster but comparable in terms of solutions quality, than other nature-inspired metaheuristics for the intrusion detection problem.
Similarly, the SA played the role of an LS improving the results generated by the ants in the ACS solving the Vehicle Routing Problem with Time Windows~\cite{a37}. 
Chen and Chien~\cite{a38} proposed a complex hybrid of four metaheuristics, namely of the Genetic Algorithm, SA, ACS, and the Particle Swarm Optimization, for solving the TSP. The SA played the role of a mutation operator in the GA part of the hybrid.
In a paper by Xi et al.~\cite{a39} the solutions generated by the ant system were later improved by the SA when solving the 3D/2D fixed-outline floor planning problem. McKendall and Shang~\cite{a40} used the SA as the LS method in one of their Hybrid Ant System algorithms for solving the dynamic facility layout problem. The resulting algorithm was able to improve some of the best known results for the problem.
Similarly, the solutions generated by the ACO were a starting point for the SA solving the problem of managing energy resources considering intensive use of electric vehicles~\cite{a41}. The combined approach produced solutions of quality better than of the SA or ACO alone.

\subsection{Sequential Ordering Problem}
\label{sec:Sequential_Ordering_Problem}

The Sequential Ordering Problem is a generalization of the Asymmetric TSP (ATSP). The goal is to find the shortest Hamiltonian path from a starting city (source node) to a destination city (final node) by going through each of the remaining cities (nodes) exactly once. Moreover, some cities have to be visited before others.
Due to precedence constraints, the problem is sometimes referred to as the Precedence Constrained Traveling Salesman problem (PCATS).
The SOP can be viewed as a scheduling problem in which many jobs have to be scheduled on a single machine. The processing times for the jobs are given along with the setup times between pairs of jobs. Also, some jobs have to be completed before others. The goal is to minimize the total makespan~\cite{a15}. 
Other real-world problems that can be modeled as an instance of the SOP include the Single Vehicle Routing Problem with pick-up and delivery constraints or the routing of a stacker crane in an automatic storage system~\cite{a14}.

An instance of the SOP can be described using a graph $G=(V, E)$ where $V$ is a set of nodes containing the starting $u_s$ and the final $v_f$ nodes, and $E = \{ (u, v) | u,v \in V, u \ne v \}$ is a set of weighted directed edges. Additionally, a precedence graph $H=(V, R)$ is given, where $R$ defines the precedence constraints, i.e. an edge $(u, v) \in R$ if node $u$ has to precede node $v$ in every \emph{feasible} solution. By definition, the starting node $u_s$ precedes every other node, i.e. $(u_s, v) \in R \; \forall v \in V \setminus \{ u_s \}$, and the final node $u_f$ has to be preceded by all other nodes, i.e. $(u, v_f) \in R \; \forall u \in V \setminus \{v_f\}$.
The precedence graph, $R$, has to be acyclic for feasible solutions to exist.

On the one hand, the precedence constraints make solving the SOP more difficult than the ATSP because the solution construction algorithms have to check for the precedence constraints. On the other hand, the precedence constraints may limit the number of feasible solutions, which can be beneficial for the exact methods, such as the branch-and-cut~\cite{a16}. All of the SOP instances from the SOPLIB2006 repository with the largest relative number of precedence constraints (60\%) were solved to optimality, whereas those in which the constraints concerned 15\% percent of all edges remained unsolved~\cite{a13}.

The SOP was introduced along with a mathematical model and exact algorithms by Escudero et al.~\cite{a19}.
Several exact approaches for solving the SOP have since been proposed. Escudero et al. applied Lagrangian Relaxation to solve the SOP, which the authors called the Relax-and-Cut algorithm~\cite{a20}.
Hern{\'a}dv{\"o}lgyi et al. proposed a branch-and-bound algorithm with the lower bounds obtained from homomorphic abstractions of the original states space~\cite{a16}. The authors solved to optimality several instances (with 40-50 vertices) from the TSPLIB repository.

Gouveia et al. proposed a cutting plane algorithm with the SOP formulations involving additional exponential-sized sets of inequalities~\cite{a17}. The authors were able to improve the best known lower bounds for many SOP instances from the TSPLIB repository and to solve to optimality instance \emph{p43.4} (the calculations took 15282 sec.).
Later, Guveia and Ruthmair solved to optimality several SOP instances from the SOPLIB2006 repository by using the branch-and-cut algorithms combined with several preprocessing methods, heuristics, and separation routines~\cite{a13}. They used a single core of an Intel Xeon E5540 or E5649 both with a 2.53GHz clock. For most of the instances the optima were found under an hour, but for 12 instances no optima were found with the time limit set to 24 hours.

The exact methods are time consuming, particularly if the size of the problem reaches a few hundred nodes, hence much of the research has been focused on heuristic algorithms for the SOP.
Guerriero and Mancini proposed a parallel roll-out heuristic in which several threads simultaneously visit different portions of the solution space and periodically exchange information about the solutions found~\cite{a21}. The algorithm was able to match the best-known solutions for most of the SOP instances from the TSPLIB repository, although its main drawback was a high computational cost.

Gambardella et al. proposed a combination of the Ant Colony System and a novel LS procedure called the \emph{SOP-3-exchange}~\cite{a11}.
The resulting algorithm, denoted as HAS-SOP, allowed to improve many best-known results for many SOP instances from the TSPLIB repository.
Monetamanni et al. added to the HAS-SOP a Heuristic Manipulation Technique which creates and adds artificial precedence constraints to the original problem~\cite{a23}. The method led to better results, particularly for large SOP instances.

A discrete Particle Swarm Optimization hybridized with the SOP-3-exchange heuristic was proposed by Anghinolfi et al.~\cite{a24}. The algorithm was able to improve many of the best results presented in~\cite{a11,a23}.
Later, Gamabrdella et al., basing their findings on an analysis of the drawbacks of the HAS-SOP algorithm, proposed an improved ACS version called the Enhanced Ant Colony System (EACS)~\cite{a22}. The two main changes were proposed. First, the construction phase of the EACS used information about the best solution found so far. Second, the LS was run only if the current solution was within 20\% of the best found solution. The EACS was able to further improve some of the best results obtained by Anghionlfi et al.~\cite{a24} and to date remains one of the most efficient methods for solving the SOP.

\section{Improving ACS Convergence with Simulated Annealing}
\label{sec:Improving_ACS_Convergence_with_Simulated_Annealing}

\label{sec:improving}


Ant colony optimization (ACO) is probably the best-known algorithm that was inspired by the foraging behavior of ants in nature.
It is a population-based meta-heuristic algorithm that is often used to solve difficult combinatorial and continuous optimization problems. In general, it does not guarantee that the optimal solution will be found, but solutions that are found are often of good enough quality (for practical use)~\cite{a9}. 

In the ACO, a number of artificial agents (ants) construct iteratively complete solutions to an optimization problem.
An ant starts with an empty solution and, in subsequent steps, extends it with components selected from the set of all available components. Each component has an associated pheromone trail and a heuristic value.
The higher the product of the pheromone concentration (value) and the heuristic value is, the higher the probability that it will be selected by the ant.

In nature, ants communicate indirectly with one another by depositing small amounts of chemical substances called \emph{pheromones}, e.g., an ant that has found a food source marks the path to the nest with small amounts of pheromone. The pheromone trail attracts other ants and leads them to the food source. The more ants that repeat the process, the higher the concentration of the pheromone trail becomes, hence the process becomes autocatalytic. The pheromone evaporates with time, so the pheromone concentration does not increase indefinitely. The ACO algorithms use artificial pheromone trails, with the pheromone concentration represented as real numbers. The set of all pheromone trails is usually called a \emph{pheromone memory} and plays a~crucial role in the performance of the ACO family of algorithms~\cite{a5, a9}.

For the TSP (and related problems) the problem is usually modeled by using a graph $G(V, E)$. An artificial ant constructs its solution starting from a randomly selected node. In subsequent steps it moves from the current node to one of the unvisited neighbor nodes by using the corresponding edge. The pheromone trails $\tau_{uv}$ are deposited on the edges, $(u,v) \in E$, of graph $G$ and together with \emph{a priori} knowledge about the problem, reflected in the heuristic values associated with each edge $\eta_{uv}$ they guide the construction process.

The Ant Colony System is an improved version of the Ant System by Dorigo et al.~\cite{a8}.
In the ACS the ant located at node $i$ selects a next node $j$ according to a \emph{pseudo-random proportional rule}~\cite{a9}:
\begin{equation}
j =\begin{cases}
\arg \max_{l \in J_k^i } [\tau_{il}] \cdot [\eta_{il}]^\beta, &
\mbox{ if }q \le q_0 \\
J,  & \mbox{ if }q > q_0,
\end{cases}
\label{eq:1}
\end{equation}

where $\eta_{il}$ is a cost associated with an edge $(i,l)$, $\tau_{il}$ is the value of the pheromone trail on edge $(i,l)$, $J_k^i$ is a set of available (candidate) nodes of ant $k$, and $q_0$ is a parameter, $0 \le q \le 1$. $J$ is a node (city) selected according to the probability distribution defined by:

\begin{equation}
   P(J | i) = \frac{ [\tau_{iJ}] \cdot [\eta_{iJ}]^\beta  }{ 
               \sum_{l \in J_k^i} [\tau_{il}] \cdot
               [\eta_{il}]^\beta }.
    \label{eq:2}
\end{equation}

The choice defined by Eq.~\ref{eq:1} depends on the value of parameter $q_0$.
If the randomly drawn number $q$ is lower than the parameter $q_0$, then the choice is greedy and the ant selects the node to which an edge leads  with the maximum product of pheromone trail $\tau_{ij}$ and heuristic $\eta_{ij}$ values. Otherwise $q \ge q_0$ and the choice is random with the probability distribution given by Eq.~\ref{eq:2}. The first case is often referred to as \emph{exploitation} of the knowledge gathered by the ants (in the pheromone memory). Usually, a value of $q_0$ close to 1 (often 0.9 and above) leads to good quality results in a shorter period of time as compared to the base ACO algorithm~\cite{a9}. Some authors even use a higher value calculated as follows: $q_0 = 1 - \frac{s}{|V|}$, where parameter $s$ is the number of nodes that should be selected randomly with the probability defined by Eq.~\ref{eq:2}~\cite{a11}.

During construction of the solutions the ants in the ACS update the values of the pheromone trails on the traversed edges. Each ant, after making a move from node $u$ to node $v$, applies a \emph{local pheromone update} rule that decreases the amount of pheromone on edge $(a,b)$ according to:
\begin{equation}
\tau_{ab} \leftarrow (1 - \psi) \cdot \tau_{ab} + \psi \cdot \tau_0 \; ,
\label{eq:3}
\end{equation}
where $\psi$ is a parameter regulating evaporation of the pheromone over time and $\tau_0$ is the initial pheromone level. The rationale behind formula (\ref{eq:3}) is that it lowers the probability of selecting the same nodes by subsequent ants, hence it increases variety in the constructed solutions.

The \emph{global pheromone update} performed after the ants have completed construction of their solutions is more important.
The update rule results in the increase in pheromone levels on trails corresponding to the best solution found so far ($\mathcal{S}_{\rm best}$) and its value by $L_{\rm best}$. For each $(u, v) \in \mathcal{S}_{\rm best}$, the pheromone changes according to the formula:
\begin{equation}
    \tau_{uv} \leftarrow (1 - \rho) \cdot \tau_{uv} + \rho \cdot \Delta \tau_{uv} \; ,
    \label{eq:global_update}
\end{equation}
where
    $\Delta \tau_{ab} = L_{\rm best}^{-1}$
and $\rho \in (0, 1)$ is a parameter regulating the strength of the pheromone increase.
The global pheromone update ensures that edges belonging to the current best solution have higher probabilities of being selected in the algorithm's subsequent iterations. The global best solution is used during the global pheromone update because it leads to slightly better solutions than the iteration best solution~\cite{a9}.

In order to further shorten the computation time of Eq.~\ref{eq:1}, the so-called \emph{candidate set} is used which consists of the nearest neighbors of the current node. The size of the candidate set is usually in the range of 10 to 25~\cite{a5,a9}. For comparison, the size $n$ of the problem is often two or three orders of magnitude larger, hence the use of candidate sets further limits exploration of the solution search space. The candidate sets are a greedy heuristic based on the observation that good quality solutions are comprised mainly of short edges. If all of the candidate set elements are already a part of the constructed solution the ant selects one of the remaining (unvisited) nodes. The candidate set is usually computed at the beginning and does not change.
Randall and Montgomery~\cite{a26} investigated the idea of dynamic candidate set updates for the TSP and the Quadratic Assignment Problem (QAP). The dynamic versions resulted in solutions of better quality but also significantly increased the computation time of the whole algorithm.

\subsection{Enhanced ACS}
\label{sec:Enhanced_ACS}

\label{sec:eacs}
The Enhanced ACS algorithm proposed by Gambardella et al. is an efficient metaheuristic for the SOP~\cite{a22}. It differs from the ACS in two ways. The first is a modified solution construction phase which is much more focused on the best solution found so far. Instead of direct application of Eq.~\ref{eq:1} an ant selects the node which follows the current node in the best solution so far (if the random number $q$ is lower than the parameter $q_0$). Only if the node is already a part of the constructed solution does the ant consider other nodes, i.e. it selects the edge with the maximum product of the pheromone and heuristic values. If $q \ge q_0$, the selection process from the ACS is used. Parameter $q_0$ usually has a value of $0.9$ or higher, hence this modification significantly speeds up the construction process although it limits the exploration capability of the EACS, and without a strong LS, it achieves results of lower quality than the ACS~\cite{a12}.

The second modification is strong integration of the solution construction phase with the LS. The LS is run only if the cost of the current solution is within 20\% of the best solution found so far. Also, the LS is initialized so that only elements of the current solution which are out of order with respect to the best solution are placed on the so-called \emph{don't push stack}. The elements of the stack are the starting points for the LS. This increases the emphasis on areas of the solution search space that are potentially unexplored.

A slightly modified version of the EACS was proposed by Ezzat~\cite{a27}. The main difference concerns the choice of the next node in the solution construction process. At first it tries to select the node $v$ which follows the current node $u$ in the best solution so far. If $v$ is already a part of the constructed solution, it selects the next node with the probability defined by Eq.~\ref{eq:3}. This change favors exploration and makes the algorithm less exploitative than the EACS but still more exploitative than the ACS.
{\R
Later Ezzat et al. adapted the EigenAnt algorithm to solve the SOP~\cite{ezzat2014}. The computational experiments showed that the proposed algorithm had performance comparable to the EACS.
}

\subsection{Simulated Annealing}
\label{sec:Simulated_Annealing}

Simulated Annealing is one of the most well-known general metaheuristic optimization methods. It was inspired by the Monte Carlo method of sampling the states of a (physical) thermodynamic system.
In the SA, a solution to the optimized problem is equivalent to a state of the thermodynamic system, and its quality corresponds to the system's current energy~\cite{a32}.
The SA works as follows: starting from an initial solution $X_0$, a sequence of solutions $(X_i), \, X \in S$ is generated, where $S$ is a set of all feasible solutions. Given a current solution $X_i$, a candidate solution $Y_i$ is generated and its cost $C(Y_i)$ is calculated. The next solution $X_{i+1}$ is selected according to:
\begin{equation}
X_{i+1} = \begin{cases}
Y_i \, , & \text{if $C(Y_i) < C(X_i)$,} \\
Y_i \, , & \text{with probability $p_i$ if $C(Y_i) \ge C(X_i)$,}\\
X_i \, , & \text{otherwise.}
\end{cases}
\end{equation}
\noindent
Probability $p_i$ is defined as 
$p_i = \exp \left( -(C(Y_i) - C(X_i)) / T_i \right)$, where $T_i > 0$ is 
called \emph{a temperature}. 
The physical analogy on which the SA is based requires that the system be kept close to a thermal equilibrium as the temperature is lowered.

The most often used cooling schedule is the \emph{exponential schedule} of the form: $T_{i+1} = \lambda T_i$, where $\lambda$ is a parameter.
In fact, the exponential cooling schedule usually lowers the temperature \emph{too fast} for the system to reach a near-equilibrium state and does not guarantee convergence to the global optimum. Nevertheless, it is useful in practice because it is easy to implement and often leads to good quality solutions if the computation time is limited~\cite{a31}. More advanced cooling schedules have been proposed; the two well-known ones are the adaptive cooling schedule by Huang et al.~\cite{a32} and the efficient cooling schedule by Lam~\cite{a30}.

\subsection{Combining ACS with Simulated Annealing}
\label{sec:Combining_ACS_with_Simulated_Annealing}

The ACS generally offers a better convergence speed than the Ant System or ACO~\cite{a5}. This stems, among others, from the more exploitative solution construction process and the global pheromone update rule that places emphasis on the best solution found so far. This usually speeds up the process of finding good quality solutions but also makes escaping local minima very difficult.
Simulated Annealing, on the other hand, offers a simple solution to escape the local minima.
We propose how to combine the ACS and SA to enhance the ACS search process while maintaining its exploitation oriented nature.
The proposed algorithm, ACS with the SA (ACS-SA in short), can be summarized as follows. The ACS search process is guided (in part) by the pheromone trail values. At the end of each iteration the global pheromone update rule increases the values of the pheromone trails corresponding to the components (edges) of the current best solution (global best). In the proposed ACS-SA algorithm the global update rule uses instead an \emph{active solution} which may not necessarily be the best solution found so far. At the end of every iteration each of the solutions generated by the ants is compared with the \emph{active solution}. If the new solution is of better quality, it replaces the current active solution. Otherwise, the new solution may still replace the active solution but with a probability defined by the Metropolis criterion known from the SA.
While the ACS is always focused on the neighborhood of the best solution found so far and can become trapped in a local optimum for a long time, the proposed ACS-SA has a greater chance of escaping the local optima by shifting focus to the solution with a higher cost.

\begin{figure}
\footnotesize
\begin{algorithm}[H]
\SetKwFor{Parfor}{for}{do in parallel}{end}

$T \leftarrow {\rm SA\_compute\_initial\_temperature({\it \gamma})}$   \tcp*[f]{  Calculate $T_0$ }

\For{ i $\leftarrow$ 1 \KwTo $\#{\it iterations}$  }{
\For{ j $\leftarrow$ 1 \KwTo $\#{\it ants}$ }{
    $route_{{\rm Ant}(j)}[1] \leftarrow \mathcal{U}\{1, \#{\it nodes}\}$  \tcp*[f]{  Start from randomly selected nodes }
}

\For(\tcp*[f]{  Build complete solutions }){ k = 2 \KwTo $\#{\it nodes}$ }{
\For{ j $\leftarrow$ 1 \KwTo $\#{\it ants}$ }{
    $route_{{\rm Ant}(j)}[k] \leftarrow $ select\_next\_node($route_{{\rm Ant}(j)}$)

    local\_pheromone\_update($route_{{\rm Ant}(j)}[k-1], route_{{\rm Ant}(j)}[k]$)
}
}
\For(\tcp*[f]{  Local update on the closing edges }){ j $\leftarrow$ 1 \KwTo $\#{\it ants}$ }{
    local\_pheromone\_update($route_{{\rm Ant}(j)}[\#{\it nodes}], route_{{\rm Ant}(j)}[1]$)
}

    $local\_best$ $\leftarrow$ select\_best($route_{{\rm Ant}(1)}, route_{{\rm Ant}(2)}, \ldots,  route_{{\rm Ant}(\#{\it ants})}$)

\If{ ${\rm Cost}(local\_best) < {\rm Cost}(global\_best)$  }{
        $global\_best$ $\leftarrow$ $local\_best$
}

SA\_select\_solution($active\_solution$, $[ route_{{\rm Ant}(1)}, route_{{\rm Ant}(2)}, \ldots,  route_{{\rm Ant}(\#{\it ants})}]$) 

\eIf( \tcp*[f]{ Allow a greedy ACS update with a small probability } ){ $\mathcal{U}(0, 1) < 0.1$ }{
    global\_pheromone\_update($global\_best$)
}
{
    global\_pheromone\_update($active\_solution$)
}

}
$T \leftarrow T \cdot \lambda$   \tcp*[f]{  Lower annealing temperature }
\end{algorithm}
\caption{Ant Colony System combined with Simulated Annealing (ACS-SA).}
\label{alg:acs-sa}
\end{figure}

{\R
Figure~\ref{alg:acs-sa} presents a pseudocode of the proposed ACS-SA algorithm. 
The major part of the algorithm does not differ from the ACS, i.e. the only differences are related to temperature initialization (line 1), the cooling schedule (line 26) and the \emph{active solution} selection process (line 19).
Inclusion of the SA into the ACS results in a more exploratory search process, but it may also lead to a prolonged examination of areas of the solution space that contain solutions of a poor quality.  
This is prevented by allowing the current \emph{global best} solution to be selected as the \emph{active solution} with a probability of 0.1 (line 20).
}
This heuristic might not be necessary if a more advanced cooling schedule is used. The present work is intended to be proof of the concept that the SA may be used to improve the convergence speed of the ACS, hence the geometric cooling schedule was adapted for its simplicity. In future work a more advanced schedule, e.g. an adaptive cooling schedule by Lam~\cite{a30}, could be applied.

Figure~\ref{alg:acs-select-solution} presents the \emph{active solution} selection procedure. The process iterates over a set of solutions built by the ants. If the cost of an ant's solution is lower than the cost of the \emph{active solution}, it replaces the \emph{active solution} (lines 3--5 in Fig.~\ref{alg:acs-select-solution}). Otherwise, the ant's solution (of a worse quality) may replace the \emph{active solution} with a probability calculated according to the Metropolis criterion from the SA (lines 6--7). As the temperature is lowered, the probability of accepting a worse solution goes down to 0 and the process becomes \emph{equivalent} to that of the ACS.

The initial temperature $T_0$ plays an important role in the SA.
In our work we applied the idea of an adaptive temperature calculation which was proposed in~\cite{a47}.
The calculation requires a sample of randomly generated solutions whose values (costs) are used to calculate the initial temperature according to:
\begin{equation}
T_0 = \frac{\overline{\Delta C} + 3 \sigma_{\Delta C}}{ \ln{ (1/\gamma) } } \; ,
\label{eq:t0}
\end{equation}
where
$\overline{\Delta C}$ is the mean of absolute differences between the costs of consecutive pairs of solutions from the sample,
$\sigma_{\Delta C}$ is the sample standard deviation
and $\gamma$ is a parameter denoting the probability of accepting a worse solution, i.e. with a higher cost.
{\R
The idea behind Eq.~\ref{eq:t0} is based on the central limit theorem which states that the mean of a large sample of independent random variables is approximately normally distributed, hence, almost all (approx. 99.7\%) absolute differences between the quality of randomly generated solutions fall in the range of
$(\overline{\Delta C} - 3 \sigma_{\Delta C}, \overline{\Delta C} + 3 \sigma_{\Delta C})$. Knowing the approximation of the highest difference in quality between a pair of solutions allows to calculate the initial temperature so that the probability of accepting a worse solution is $\gamma$.
}

Although the temperature initialization requires additional computations, it does not increase the asymptotic complexity of the ACS algorithm. In our experiments a sample of 1000 random solutions was used due to a negligible additional cost; however, a much smaller number could also be acceptable.

\begin{figure}
\footnotesize
\begin{procedure}[H]

{\bf Procedure} SA\_select\_solution($active\_solution$,$solutions$)

\For{ i $\leftarrow$ 1 \KwTo $\#{\it ants}$ }{

    \eIf{ {\rm Cost}($solution[i]$) $ < $ {\rm Cost}($active\_solution$)  }{
            $active\_solution \leftarrow solution[i]$
    }
    {
        $\Delta C \leftarrow {\rm Cost}(solution[i]) - {\rm Cost}(active\_solution)$
        
        \If{ $\mathcal{U}(0,1) < e^{-\Delta C/T}$ }{
                $active\_solution \leftarrow solution[i]$
        }
    }

}
\end{procedure}
\caption{Simulated Annealing-related procedure to update the $active\_solution$.}
\label{alg:acs-select-solution}
\end{figure}

\subsection{Combining Enhanced ACS with the SA}
\label{sec:Combining_Enhanced_ACS_with_the_SA}

As described in Sec.~\ref{sec:eacs}, the EACS differs only slightly from the ACS. The differences are minor and concern the solution construction process and the LS application, hence it is straightforward to apply exactly the same ideas to incorporate the SA into the EACS as in the proposed ACS-SA algorithm.
Due to its more (i.e. relative to the ACS) exploitative behavior, the EACS is even more susceptible to getting trapped in the local minima, hence it should also benefit from the SA component. The resulting algorithm will henceforth be denoted as the EACS-SA.

\section{Efficient Local Search for the SOP}
\label{sec:Efficient_Local_Search_for_the_SOP}

\label{sec:local-search}

Even though the ACS, MMAS and related algorithms perform competitively to other nature inspired metaheuristics their 
convergence can still be improved with a problem-specific local search~\cite{a9}.
When combined with the LS, the ACS is responsible for finding a candidate solution, while the aim of the LS is to improve it by performing small changes leading to a neighboring solution of a better quality.
In this section we start with a description of the state-of-the-art LS heuristic for the SOP and later propose a modified version which incorporates the SA component.

\subsection{SOP-3-exchange}
\label{sec:SOP_3_exchange}

Gambardella et al.~\cite{a11} proposed an efficient LS heuristic for the SOP called the \emph{SOP-3-exchange}. It adapts the 3-opt heuristic known from the TSP to the SOP without an increase in algorithm time complexity. The SOP-3-exchange belongs to the family of edge-exchange procedures, in which a new solution is generated by replacing $k$ existing edges with another set of $k$ edges for which the cost of the solution is lower. This operation is usually called $k$-exchange, and the value of $k$ can be fixed (typically 2 or 3) or can vary as in the Lin-Kernighan heuristic~\cite{a28}. Starting from the initial solution and applying the $k$-exchange iteratively until no further improving exchange exists leads to a \emph{k-optimal} solution. This process requires in the worst-case scenario, $O(n^k)$ time.

During a $k$-exchange procedure $k$ existing edges are removed producing $k$ disjoined paths which are then reconnected with $k$ new edges. In some cases, reconnection of the paths requires that some of them be reversed, e.g. in the case of a 2-opt move and a closed path $<0, \ldots, i-1, i, i+1, \ldots, h-1, h, h+1, \ldots, n-1>$; there are two possible ways to reconnect the subpaths after removal of the $(i, i+1)$ and $(h, h+1)$ edges, namely $<\ldots, h+1, i+1, \ldots, h-1, h, i, i-1, \ldots>$ and $<\ldots, i-1, i, h, h-1, \ldots, i+1, h+1, \ldots>$; both require a reversal of the subpath.  The reversal, however, is problematic for the SOP because the distances between the nodes are asymmetric, hence the length of the path after the reversal should be recalculated what requires $O(n)$ time. Because the cost of a $k$-opt move should be calculated in a constant time an efficient implementation of the $k$-opt heuristic for the SOP should be restricted only to \emph{path-preserving} exchanges~\cite{a11}. 

\begin{figure}[p]
    \centering
    \includegraphics[width=0.8\textwidth]{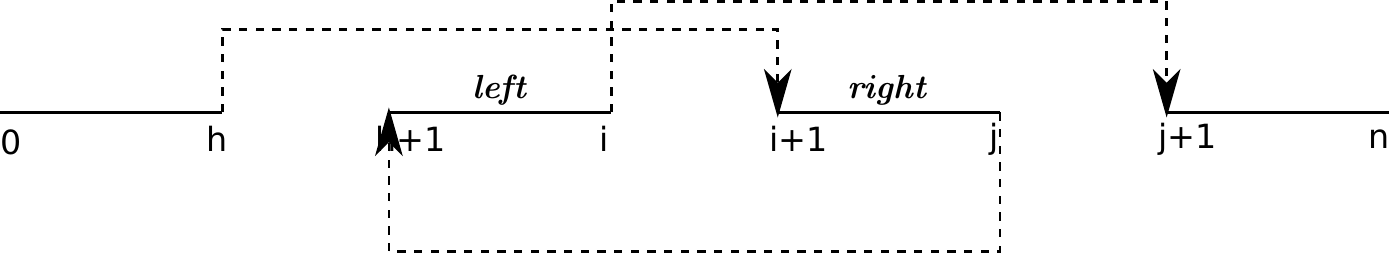}
    \caption{Path-preserving-3-exchange for the SOP}
    \label{fig:sop3-exchange}
\end{figure}

The smallest $k$ that allows a path-preserving exchange is $k=3$, denoted as the \emph{path-preserving-3-exchange} shown in Fig.~\ref{fig:sop3-exchange}. 
By removing the $(h, h+1)$, $(i, i+1)$ and $(j, j+1)$ edges and adding $(h, i+1)$, $(j, h+1)$ and $(i, j+1)$ edges the two neighboring subpaths are swapped, thus preserving the relative order of the elements.
After performing the path-preserving-3-exchange one would still need to verify if the precedence constraints for the two subpaths are preserved. This requires $O(n^2)$ time in the general case but can be avoided as in the method proposed by Gambardella et al.~\cite{a11}.
There are two necessary procedures to reduce the computation time. The first requires keeping the \emph{lexicographic} order while searching for the path-preserving-3-exchange. The second, is the use of a \emph{labeling} method.

Figure~\ref{fig:sop3-exchange} shows how the route changes when applying the path-preserving-3-exchange. The tree indices $h$, $i$, and $j$ ($h < i < j$) define two sub-paths in the route: left $(h+1, i)$ and right $(i+1, j)$. The subpaths are swapped as a result of performing the exchange, i.e. the \emph{right} path comes before the \emph{left} path. This can only happen if there are no precedence constraints between the considered node and the nodes in the left path.

The path-preserving-3-exchange as proposed by Gambardella et al.~\cite{a11} consists of \emph{forward} and \emph{backward} searches for feasible path-preserving-3-exchanges.
The \emph{forward} search involves incrementing $j$ iteratively, thus increasing the length of the right path by one. This requires that only the precedence constraints be checked between the elements of the left path and the node considered for inclusion into the right path. Eventually, a precedence constraint is hit and the procedure is repeated with the left path being extended with a single element (by incrementing $i$) and the right path set to a single element, i.e. $(j)$, $j = i+1$. 
After all of the possibilities are exhausted $h$ is incremented and the process repeats for all possible $i$ and $j$ values ($i < j < n$).
This leads to $O(n^3)$ possible pairs of subpaths each requiring $O(n)$ constraints verification, hence a total complexity of $O(n^4)$.

The cost of constraints verification can be reduced to $O(1)$ due to the \emph{labeling procedure}. The procedure works as follows. Each time the left subpath is extended with a new node $u$ (during the sop-3-exchange), \emph{mark($v$)} is set to \textit{count} for every node $v$ for which there exists a precedence constraint between $u$ and $v$.
The \textit{count} is a variable initially set to 0 and incremented each time the left path grows, i.e. $h$ is incremented.
Thanks to the procedure, each time the right path is extended with a node $x$ one needs only to check the value of $\textrm{mark}(x)$. If the value equals the \textit{count}, then the node at index $j$ (in the right path) has to be visited after the nodes in the left path, hence the two paths cannot be swapped. This reduces the complexity of the whole search for a feasible path-preserving-3-exchange to $O(n^3)$, which is asymptotically equal to the complexity of a 3-opt heuristic used to solve the TSP.

The forward search for the path-preserving-3-exchange considers only exchanges defined by indices $i$, $j$ and $h$ such that $0 < h < i < j < n$, where $n$ is the number of nodes. The \emph{backward} search is analogous to the forward search but the left and right paths are expanded in the direction of decreasing indices, i.e. the left path "moves" from the end of the sequence to the beginning.

Summarizing, the time complexity of finding a single profitable 
\emph{path-preserving-3-exchange} using the described procedure is $O(n^3)$.
This is still expensive as the procedure is applied (to a single solution) in a loop until no further improving move is found
and it has to be repeated for the subsequent solutions.
Gambardella et al.~\cite{a11} proposed two additional changes to reduce the algorithm's computation time. The first is to limit the search to only a subset of all potential moves. By default the SOP-3-exchange for each index $h$ considers all valid $i$ and $j$ indices. Assuming most of the changes will involve relatively short paths, the $i$ values can be restricted to $h+1, h+2, h+3$ for the forward procedure and $h-1, h-2, h-3$ for the backward procedure. This version was named \emph{OR-exchange}~\cite{a11}.
The second change involves the use of two additional heuristics, i.e. \emph{don't look bits} and \emph{don't push stack}.
\emph{Don't look bits} is a data structure that was proposed by Bentley~\cite{a29} which works as follows. A bit is associated with each node of the solution. At the beginning all the bits are turned off and are turned on when the SOP-3-exchange starts looking for a profitable exchange originating from the node. If the \emph{don't look bit} is turned on, the corresponding node is ignored by the subsequent SOP-3-exchange searches until the node is involved in a profitable \emph{path-preserving-3-exchange}. Then all six pivot nodes ($h, h+1, i, i+1, j, j+1$) are turned off. Use of the \emph{don't look bits} aims to focus the search on the changing parts of the solution.
The purpose behind the use of the \emph{don't push stack} is similar -- it contains nodes $h$ to be selected as a starting point of a \emph{path-preserving-3-exchange}. At the beginning the stack is initialized with all of the nodes. During the search a node, $h$, is removed from the stack and if the feasible move originating from node $h$ is found the six nodes involved in the exchange are pushed onto the stack (if they do not belong to it already).
An additional benefit of using the \emph{don't push stack} is that the linear order in which the nodes are considered during the search for a profitable \emph{path-preserving-3-exchange} is broken.

\begin{figure}
\footnotesize

\begin{algorithm}[H]
\SetKwInOut{Input}{Procedure}
\Input{SOP\_3\_exchange}

\SetKwInOut{Input}{Input}
\SetKwInOut{Output}{Output}

\Input{$route[0..n]$ \tcp*[f]{ A solution to improve } }
\Input{$best\_route[0..n]$ \tcp*[f]{ The current best solution } }

$S$ $\leftarrow$ init\_dont\_push\_stack($route$, $best\_route$)

\While{ $S \ne \emptyset$ }{
    $h$ $\leftarrow$ pop\_stack($S$)

    \If{ {\rm forward\_SOP\_3\_exchange}(h,route) $\ne$ {\rm \bf True} }{
            backward\_SOP\_3\_exchange($h$,$route$)
    }
}
\end{algorithm}
\caption{Pseudocode of the  SOP-3-exchange procedure.}
\label{alg:sop-3-exchange}
\end{figure}

\begin{figure}
\footnotesize
\begin{procedure}[H]
\SetKwInOut{Input}{Function}
\Input{forward\_SOP\_3\_exchange}
\SetKwFor{ForEach}{foreach}{do}{end}
\SetKwInOut{Input}{Input}
\SetKwInOut{Output}{Output}
\Input{$h \in \{0,1,\ldots,n-2\}$  \tcp*[f]{ An index of the SOP-3-exchange start element } }
\Input{$route[0..n]$  \tcp*[f]{ A SOP solution being improved }}

\Output{\textbf{True} if an acceptable change was found, \textbf{False} otherwise}

$succeded$ $\leftarrow$ \textbf{False}

$label[0..n]$ $\leftarrow$ $\emptyset$

\For{ i $\leftarrow$ h+1 \KwTo n-1 }{
\ForEach(\tcp*[f]{  Labelling procedure }){ node $u$ preceded by route[i] }{
$label[u]$ = $h$  \tcp*[f]{ Node $u$ comes after route[i] }
}

   $best\_delta$ $\leftarrow$ 0

   $best\_j$ $\leftarrow$ $\emptyset$

   $j$ $\leftarrow$ $i+1$

\While{ $label[j] \ne h$ }{
       $delta$ $\leftarrow$ calculate\_cost\_change(h,i,j)

\If{ {\rm is\_move\_accepted}(delta,best\_delta)  }{
           $best\_j$ $\leftarrow$ $j$

           $best\_delta$ $\leftarrow$ $delta$
}

       $j$ $\leftarrow$ $j+1$
}

\If{ $best\_j \ne \emptyset$  }{
perform\_exchange($h,i,best\_j$)  \tcp*[f]{  Exchange paths $(h+1,i)$ and $(i+1,best\_j)$ }

       $succeded$ $\leftarrow$ \textbf{True}

       \textbf{break}
}
}
\end{procedure}
\caption{Forward part of the \emph{SOP-3-exchange} algorithm that searches for a feasible move in the \emph{forward} direction
starting at a given node $h$.
The search in the \emph{backward} direction is analogous with indices $i$ and $j$ being decreased instead of increasing.}
\label{alg:sop-3-exchange-forward}
\end{figure}

Figure~\ref{alg:sop-3-exchange} presents a pseudocode of the SOP-3-exchange algorithm.
It starts with an initialization of the \emph{don't push stack} and repeatedly searches for a feasible move.
First, it searches in the forward direction (line 4) and if it fails, the backward search is applied (line 5).
The pseudocode of the search in the forward direction is shown in Fig.~\ref{alg:sop-3-exchange-forward} (the search in the backward direction is analogous).
It starts with a given index $h$ that denotes the starting point of a possible \emph{path-preserving-3-exchange}
and searches for the remaining two points, denoted by indices $i$ and $j$. The labeling procedure is applied
incrementally (lines 4--6). The function \emph{is\_move\_accepted} in line 12 simply checks if the proposed decrease in the solution value is better than the current best, but it can be replaced by a more advanced criterion as will be shown later.

\subsection{Improving SOP-3-exchange Efficiency with SA}
\label{sec:Improving_SOP_3_exchange_Efficiency_with_SA}

The SOP-3-exchange LS is efficient in improving solutions generated by the ants; however, the improvement process is greedy and only better (downhill) moves are accepted. It makes it possible to reach a local optimum quickly; however, it also makes it unable to find any better solution that would require making at least one uphill move. 
Similarly to our idea of incorporating the SA into the ACS and EACS algorithms, we propose to include the SA decision process into the SOP-3-exchange in order to make it more explorative.
The proposed modification is easy to implement as it only requires to modify the greedy condition as to whether to accept a given subpath exchange in the forward search for a path-preserving-3-exchange (line 12 in Fig.~\ref{alg:sop-3-exchange-forward}) (analogously in the backward search).
The pseudocode of the proposed modification is shown in Fig.~\ref{alg:accept-move-sa}. The decision whether to accept the proposed move (subpath exchange) is made based on the change (decrease) in the solution value and the value of the best move found so far. If the proposed move is better than the current best, it is always accepted. Otherwise, if it results in the same decrease of the solution length then it is accepted with a probability of 10\% (lines 4--5 in Fig.~\ref{alg:accept-move-sa}). It allows to accept moves which do not change the solution value but which result in a different relative order of the solution nodes.
Finally, if the proposed move is worse than the best move found so far, it is accepted with the probability calculated using the Metropolis criterion, as in the SA.

Similarly to the ACS-SA, there are two parameters related to the SA component of the proposed \emph{SOP-3-exchange-SA} algorithm, namely $\lambda_{\rm LS}$ and $\gamma_{\rm LS}$.
The former is used in the geometric cooling schedule to lower the temperature $T_{\rm LS}$,
while $\gamma_{\rm LS}$ is related to an initial probability of accepting a worse move.
There is, however, a slight difference in the temperature initialization relative to the ACS-SA.
In the ACS-SA the initial temperature is calculated based on a sample of differences in the quality (length) of the randomly generated solutions, just before the main computations.
In the SOP-3-exchange-SA the sample comes from the values of the differences in the solution quality (delta values) resulting from the subsequent path-preserving-3-exchanges considered during the initial runs of the SOP-3-exchange-SA (lines 11--16 in Fig.~\ref{alg:accept-move-sa}).
In other words, there is no dedicated temperature initialization phase in the SOP-3-exchange-SA and the sample of delta values is collected on the run in order not to slow down the whole algorithm.
After the sample of $10^5$ (a value found experimentally) is collected, the initial value of temperature $T_{\rm LS}$ is calculated, and in subsequent invocations of the SOP-3-exchange-SA the temperature is reset to this initial value without recalculating.


\begin{figure}
\footnotesize
\begin{procedure}[H]
\SetKwInOut{Input}{Function}
\Input{is\_move\_accepted\_SA}
\SetKwFor{Parfor}{for}{do in parallel}{end}
\SetKwInput{Input}{Input}
\SetKwInput{Output}{Output}
\SetKw{KwAnd}{and}
\SetKw{True}{True}

\Input{{\it change} \tcp*[f]{A decrease in a solution length if the move is applied}}

\Input{{\it best\_change} \tcp*[f]{The value of the best move found so far}}

\Output{{\bf True} if the move value should be accepted, {\bf False}}

$\Delta \leftarrow change - best\_change$ \tcp*[f]{$\Delta$ is used to compare the proposed move with the current best }

$accept$ $\leftarrow$ \textbf{False}

\uIf{ $\Delta > 0$  }{
    $accept$ $\leftarrow$ {\bf True}  \tcp*[f]{  Always accept a better move }
}
\uElseIf( \tcp*[f]{ Move does not change solution value } ){$\Delta = 0$ \KwAnd $\mathcal{U}(0,1) < 0.1$ }{
      $accept \leftarrow $  \True  \tcp*[f]{ but changes relative order of the nodes }
}
\uElse( \tcp*[f]{ Worse move, apply the Simulated Annealing criterion } ){
    \eIf{ temperature $T_{\rm LS}$ was initialized }{

        set $accept$ to {\bf True} with prob. $e^{-\Delta/T_{\rm LS}}$

        $T_{\rm LS}$ $\leftarrow$ $\lambda_{\rm LS} T_{\rm LS}$  \tcp*[f]{  Lower the temperature }

    }( {\it temperature was not yet initialized} ){

         $D$ $\leftarrow$ $D \cup \Delta$  \tcp*[f]{ Collect a sample of $\Delta$ values }

         \If( \tcp*[f]{ Calc. the initial temp. if the collected sample is large enough} ){$|D| \ge 10^5$}{
             $T_{\rm LS} \leftarrow$ calc\_initial\_temperature($D$)
         }

    }
}
\end{procedure}
\caption{Simulated Annealing-related criterion used to accept a solution change (move) in the SOP-3-exchange-SA local search algorithm.}
\label{alg:accept-move-sa}
\end{figure}

\section{Computational Experiments}
\label{sec:Computational_Experiments}

\label{sec:experiments}

A series of computational experiments was conducted in order to evaluate the performance of the proposed algorithms.
In the first part of the experiments we focused on the efficiency of the ACS-SA and EACS-SA used alone, i.e. without the problem-specific LS. 
In the second part the focus was placed on the efficiency of the algorithms coupled with the SOP-3-exchange and SOP-3-exchange-SA LS heuristics. 

The ACS and EACS require that a number of parameters be set. Based on preliminary computations and suggestions from the literature we used the following settings in our experiments: 
number of ants, $m=10$; $\beta = 0.5$; $\psi = 0.01$ and $\rho = 0.1$, and local and global pheromone evaporation ratio, respectively; $q_0 = \frac{n-20}{n}$, where $n$ is the size of the problem.
The computations were repeated 30 times for each configuration of the parameter values and the problem instance.
The computations were conducted on a machine equipped with a Xeon E5-2680v3 12
core CPU clocked at 2.5GHz, although a single core was used per run. All
algorithms were implemented in C++ and compiled with the GNU compiler with the
-Ofast switch\footnote{The source code is available at https://github.com/RSkinderowicz/AntColonySystemSA}.

\subsection{ACS-SA Parameter Tuning}
\label{sec:ACS_SA_Parameter_Tuning}

\label{sec:acs-sa-parameters-tuning}
The first part of the experiments was focused on the behavior of the ACS-SA algorithm depending on the values of the SA-related parameters.
The proposed ACS-SA algorithm uses a simple exponential cooling schedule 
$T_k = T_0 \cdot \lambda^k$, where $\lambda < 1$ is the cooling factor and $T_0$ is the initial temperature.
Although the exponential cooling schedule does not guarantee convergence to a global optimum, it has the advantage of being easy to implement and often performs well in practice~\cite{a43}.
Preliminary computations showed that the most important factor for the performance of the ACS-SA was the $\lambda$ parameter which directly influences the speed of the SA convergence. The best performance was observed for $\lambda \ge 0.999$, for which the probability of accepting worse quality solutions and, hence, escaping local minima remained high for a relatively long time.
It is not without significance that the algorithm was run for $10^5$ iterations, and for shorter/longer runs a smaller/higher $\lambda$ value could prove better. In fact, the $\lambda$ value could be calculated based on the total number of iterations if used in practice~\cite{a43}.
A number of "promising" values was selected for a more thorough investigation, namely $\lambda \in \{ 0.999, 0.9995, 0.9999 \}$.
The initial temperature $T_0$ was calculated for each problem instance during the initialization phase so that the probability of accepting a worse solution (an uphill move) at the beginning was approx. equal to the specified probability $\gamma$ (a parameter, independent of a problem instance). The mean difference between the successive solutions was estimated based on a sample of randomly generated solutions. In our experiments we considered $\gamma \in \{ 0.1, 0.5, 0.9 \}$ leading to a total of 9 combinations of $\lambda$ and $\gamma$.
The algorithm was run for a total of 14 SOP instances from the TSPLIB repository, namely: \emph{ft53.1, ft53.4, ft53.3, ft53.2, ft70.4, ft70.3, ft70.2, ft70.1, prob.100, kro124p.4, kro124p.3, kro124p.2} and \emph{kro124p.1}.

We used statistical tests to verify if the results for the various values of parameters differed significantly.
The proposed experimental design can be viewed as a \emph{two-way} (two-factor)
layout in which the main factor is the combination of $\lambda$ and $\gamma$ values, while the second factor (also called a blocking factor) is the problem instance (13 instances in our case)~\cite{a44}. More specifically, the design can be described as a \emph{randomized block design} with an equal number of replications per treatment-block combination. A suitable non-parametric (distribution-free) statistical test was proposed by Mack and Skillings and is an equivalent of a parametric two-way ANOVA~\cite{a45}. The null hypothesis, $H_0$, which is of our interest here is that of no differences in the medians (of the solution quality) for algorithms with various $\lambda$ and $\gamma$ values considered here (a total of 9 combinations). 
Rejecting the null hypothesis would mean that the different values of the $\lambda$ and $\gamma$ parameters lead various performance of the ACS-SA.
The test requires that the Mack-Skillings statistic be computed (\textit{MS}) which is then compared with a critical value $ms_\alpha$ at the $\alpha$ level of significance ($\alpha=0.05$ in our case)~\cite{a44}. The null hypothesis $H_0$ is rejected if $\textit{MS} \ge \textit{ms}_\alpha$.
In our case $\textit{MS} \approx 72.68$ while the critical value $\textit{ms}_{0.05} \approx 15.23$, hence $H_0$ was rejected, providing rather strong evidence that the values of $\lambda$ and $\gamma$ have a significant impact on the quality of the results generated by the ACS-SA. This is an expected result because the value of $\lambda$ should have a strong effect on the search trajectory of the SA.

\begin{table}[]
\centering
\caption{Table containing the $p$-values of the post-hoc pairwise comparison between the results of the ACS-SA with various
$(\lambda, \gamma)$ values (shown in the second row) according to the non-parametric, two-sided multiple comparison 
procedure by Mack and Skillings at $\alpha=0.05$~\cite{a44}. The test itself corrects for the Type I family-wise error.
The +/- symbol after a value denotes that the configuration in a row was significantly better/worse than the configuration in a column.
}
\label{tab:acs-sa-p-table}
\resizebox{\textwidth}{!}{%
\begin{tabular}{@{}llllllllll@{}}
\toprule
  & \multicolumn{1}{c}{A} & \multicolumn{1}{c}{B} & \multicolumn{1}{c}{C} & \multicolumn{1}{c}{D} & \multicolumn{1}{c}{E} & \multicolumn{1}{c}{F} & \multicolumn{1}{c}{G} & \multicolumn{1}{c}{H} & \multicolumn{1}{c}{I} \\
  & (0.999, 0.1)          & (0.999, 0.5)          & (0.999, 0.9)          & (0.9995, 0.1)         & (0.9995, 0.5)         & (0.9995, 0.9)         & (0.9999, 0.1)         & (0.9999, 0.5)         & (0.9999, 0.9)         \\ \midrule
A &   --                     & 0.9970                & 1.0                   & 0.3863                & 0.9740                & 0.9595                & 0.0004-               & 0.4397                & 0.0205+               \\
B & 0.9970                &   --                     & 0.9765                & 0.8856                & 1.0                   & 1.0                   & 0.0093-               & 0.9155                & 0.0010+               \\
C & 1.0                   & 0.9765                &   --                     & 0.2235                & 0.9019                & 0.8678                & 0.0001-               & 0.2639                & 0.0509                \\
D & 0.3863                & 0.8856                & 0.2235                &   --                     & 0.9703                & 0.9817                & 0.4191.0              & 1.0                   & $<0.0001+$               \\
E & 0.9740                & 1.0                   & 0.9019                & 0.9703                &   --                     & 1.0                   & 0.0265-               & 0.9813                & 0.0002+               \\
F & 0.9595                & 1.0                   & 0.8678                & 0.9817                & 1.0                   &   --                     & 0.0346-               & 0.9891.0              & 0.0002+               \\
G & 0.0004+               & 0.0093+               & 0.0001+               & 0.4191.0              & 0.0265+               & 0.0346+               &   --                     & 0.3668                & $<0.0001+$               \\
H & 0.4397                & 0.9155                & 0.2639                & 1.0                   & 0.9813                & 0.9891.0              & 0.3668                &   --                     & $<0.0001+$               \\
I & 0.0205-               & 0.0010-               & 0.0509                & $<0.0001-$              & 0.0002-               & 0.0002-               & $<0.0001-$              & $<0.0001-$              &   --                     \\ \bottomrule
\end{tabular}%
}
\end{table}

\begin{figure}[!ht]
    \subfloat[ACS-SA convergence for the \emph{kro124p.2} instance.\label{fig:acs-sa-convergence:a}]{%
      \includegraphics{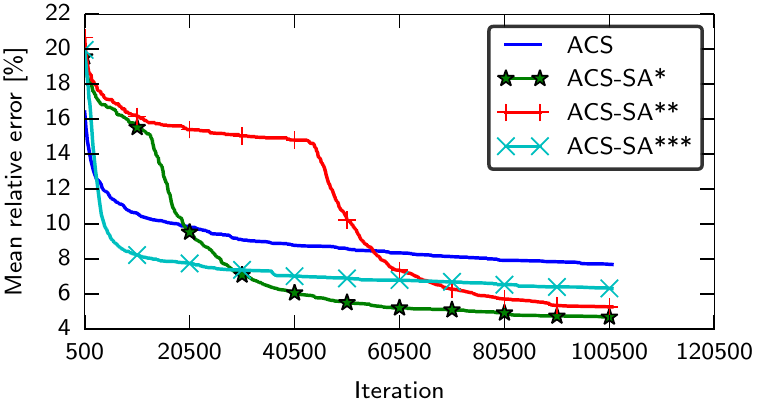}
    }
    \hfill
    \subfloat[ACS-SA convergence for the \emph{ft53.4} instance.\label{fig:acs-sa-convergence:b}]{%
      \includegraphics{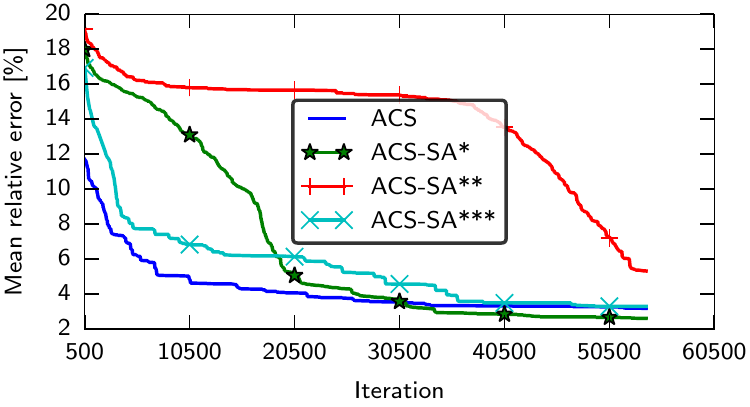}
    }
    \caption{
    Convergence of the ACS and ACS-SA for the two SOP instances. The ACS-SA was run with different
    $(\lambda, \gamma)$ values, namely: ACS-SA* -- $(0.9999, 0.1)$, ACS-SA** -- $(0.9999, 0.9)$
    and ACS-SA*** -- $(0.999, 0.1)$.
    The first 500 iterations were skipped for clarity.
    }
    \label{fig:acs-sa-convergence}
\end{figure}

After the rejection of $H_0$, we can apply a post-hoc test to compare  
the individual pairs of algorithm results obtained for respective pairs of $\lambda$ and $\gamma$ values. A suitable asymptotically distribution-free, two-sided, multiple comparison procedure using within-block ranks was proposed by Mack and Skillings~\cite{a44,a45}. Table~\ref{tab:acs-sa-p-table} contains the final $p$-values of the pairwise comparison.
As can be observed, in most cases there were no significant differences between the results of the ACS-SA with the various $\lambda$ and $\gamma$ values.
The only exception was configuration $\lambda=0.9999$ and $\gamma = 0.1$, for which the results were significantly better 6 out of 8 times.
On the other hand, configuration $\lambda = 0.9999$ and $\gamma=0.9$ was worse 7 out of 8 times. This shows that the SA component of the ACS-SA has the strongest influence if the temperature is decreased slowly. It is important to properly adjust the initial probability $\gamma$ of accepting a worse quality solution and, hence, the initial temperature $T_0$. If the probability is high, the algorithm easily accepts inferior solutions, particularly at the beginning of the computations, and drifts away from the good quality solutions. It is worth emphasizing that these observations are valid for the computation budget (time) used in the experiments; greatly increasing the computation time could show even better convergence for higher $\gamma$ values. Figure~\ref{fig:acs-sa-convergence} shows the convergence plots for the ACS-SA with various $\lambda$ and $\gamma$ levels: for $\lambda = 0.999$ the temperature drops relatively quickly and convergence of the ACS-SA resembles that of the ACS. For $\lambda = 0.9999$ the temperature drops more slowly and the algorithm has a greater chance of escaping the local minima for a longer period of time. By increasing the initial temperature (as for $\gamma = 0.9$) we can extend 
the initial "free wandering" phase at the expense of slower convergence.

\subsection{ACS-SA and EACS-SA Performance}
\label{sec:ACS_SA_and_EACS_SA_Performance}

The first experiment showed that the SA component indeed had a significant impact on ACS-SA search convergence.
In the subsequent experiment we focused on a comparison between the ACS-SA relative to the ACS. 
We also considered the EACS and the EACS combined with the SA (EACS-SA).
Both the ACS-SA and the EACS-SA were run with $\lambda = 0.9999$ and $\gamma = 0.1$, chosen based on the previous experiment.
To make the comparison fair, all of the algorithms were run with a time limit of 60 seconds of CPU time.
Although the limit was relatively low it was sufficient to detect differences in the performance of the algorithms.
A total of 20 SOP instances of varying size were selected from the TSPLIB repository.

\begin{figure}[!ht]
    \centering
    \includegraphics{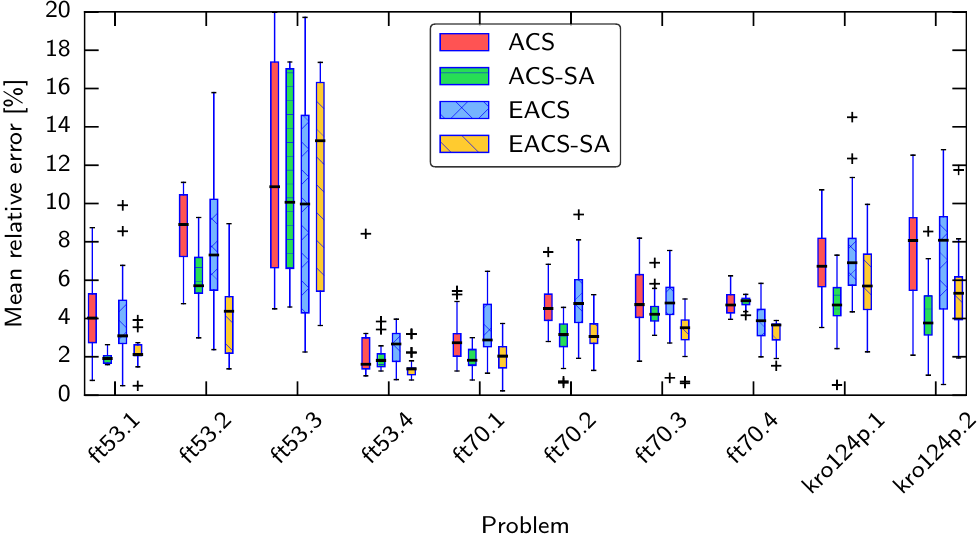}
    \caption{
    Boxplot of the mean solution error for the ACS, ACS-SA, EACS and EACS-SA algorithms 
    for the (smaller) SOP instances from the TSPLIB repository.
    }
    \label{fig:sop-cmp-a}
\end{figure}

\begin{figure}[!ht]
    \centering
    \includegraphics{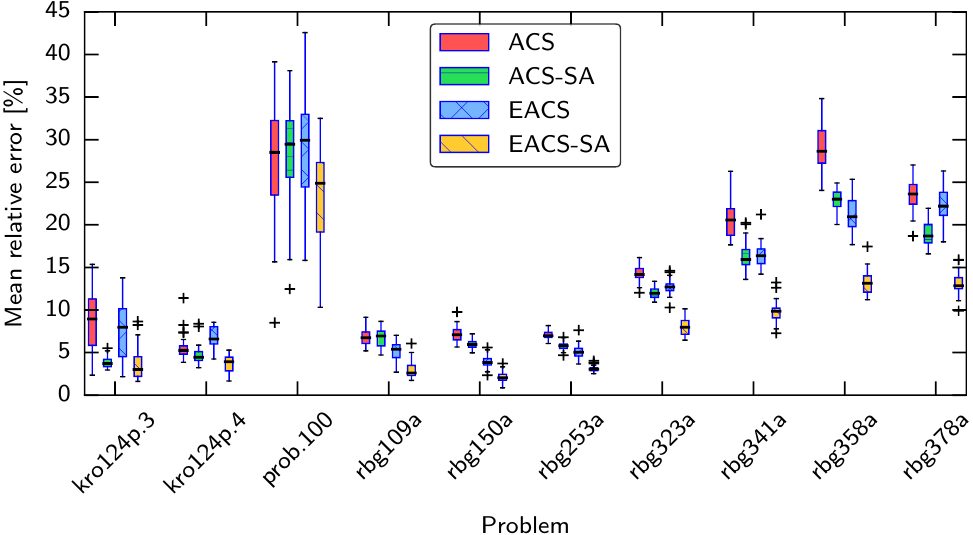}
    \caption{
    Boxplot of the mean solution error for the ACS, ACS-SA, EACS and EACS-SA algorithms 
    for the (larger) SOP instances from the TSPLIB repository.
    }
    \label{fig:sop-cmp-b}
\end{figure}

The boxplots of the mean solution error are shown in Fig.~\ref{fig:sop-cmp-a} and Fig.~\ref{fig:sop-cmp-b}.
The differences between the quality of the solutions generated by the algorithms are clearly visible. For the smaller instances, performance of the ACS and EACS was relatively similar and, in most cases, worse than that of the ACS-SA and EACS-SA, respectively.
The differences became more distinct for larger instances (up to 380 nodes), for which the EACS outperformed the ACS in most cases. The ACS-SA generally beat the ACS but even better performance was achieved by the EACS-SA version, particularly for the largest instances.
The results were compared statistically to make the comparison more complete.
For each problem instance, the Kruskal-Wallis non-parametric one-way analysis of variance test (an extension of the Mann-Whitney U test) with $\alpha=0.05$ was applied to check the hypothesis that the results of the four algorithms came from the same distribution. 
The hypothesis was rejected in 19 out of 20 cases meaning that the results of the algorithms differed significantly.
In such cases a post-hoc test was applied to compare all pairs of results. 
For this purpose the Bonferroni-Dunn test was employed with a family-wise Type I error correction ($\alpha_{FW}=0.05$)~\cite{a46}.
The results are summarized in Tab.~\ref{tab:sop-cmp}. For each pair of algorithms, only the final verdict is shown with a letter indicating the algorithm that achieved significantly better results than the others. 
The ACS-SA outperformed the ACS in 12 cases, while never generating worse results. The SA component is actually more beneficial than the exploitation-oriented heuristics introduced in the EACS which generated significantly better quality results only in 7 cases as compared to the ACS.
The EACS-SA outperformed both the EACS and the ACS in 16 out of 20 cases. The greatest difference could be observed particularly for the larger SOP instances.

\begin{table}[]
\centering
\caption{
Summary of results obtained by the ACS, ACS-SA, EACS and EACS-SA algorithms on a set of 20 SOP instances from the TSPLIB repository.
The left-most part of the table contains the mean solution lengths along with the standard deviations shown in the braces. The last 6 columns contain a 
summary of the statistical comparison between the respective pairs of algorithms
according to the two-sided, non-parametric Bonferroni-Dunn test with a family-wise Type I error correction ($\alpha_{FW}=0.05$). The capital letter indicates the algorithm which obtained significantly better results than the others. Hyphens denote that there were no significant differences between the results of the respective algorithms.
}
\label{tab:sop-cmp}
\resizebox{\textwidth}{!}{%
\begin{tabular}{@{}lllllrrrrrr@{}}
\toprule
Problem   & ACS (A)        & ACS-SA (B)    & EACS (C)       & EACS-SA (D)   & \multicolumn{1}{l}{A vs B} & \multicolumn{1}{l}{A vs C} & \multicolumn{1}{l}{A vs D} & \multicolumn{1}{l}{B vs C} & \multicolumn{1}{l}{B vs D} & \multicolumn{1}{l}{C vs D} \\ \midrule
ft53.1    & 7857 (161.8)   & 7673 (18.5)   & 7818 (162.9)   & 7702 (46.1)   & B                          & -                          & D                          & B                          & -                          & D                          \\
ft53.2    & 8713 (159.5)   & 8522 (107.0)  & 8647 (256.8)   & 8348 (156.6)  & B                          & -                          & D                          & -                          & D                          & D                          \\
ft53.3    & 11506 (578.9)  & 11417 (506.4) & 11271 (605.5)  & 11418 (544.0) & -                          & -                          & -                          & -                          & -                          & -                          \\
ft53.4    & 14744 (201.8)  & 14704 (81.7)  & 14779 (128.2)  & 14639 (101.1) & -                          & -                          & D                          & -                          & D                          & D                          \\
ft70.1    & 40437 (458.0)  & 40054 (223.6) & 40692 (588.6)  & 40150 (345.5) & B                          & -                          & -                          & B                          & -                          & D                          \\
ft70.2    & 42263 (454.5)  & 41629 (409.0) & 42396 (664.4)  & 41710 (355.0) & B                          & -                          & D                          & B                          & -                          & D                          \\
ft70.3    & 44674 (667.0)  & 44388 (333.3) & 44589 (570.0)  & 43946 (436.6) & -                          & -                          & D                          & -                          & D                          & D                          \\
ft70.4    & 56098 (325.5)  & 56146 (127.0) & 55593 (564.1)  & 55305 (362.9) & -                          & C                          & D                          & C                          & D                          & -                          \\
kro124p.1 & 42166 (757.8)  & 41313 (572.8) & 42324 (988.9)  & 41768 (731.4) & B                          & -                          & -                          & B                          & -                          & -                          \\
kro124p.2 & 44548 (1113.2) & 43049 (764.4) & 44270 (1318.3) & 43529 (809.9) & B                          & -                          & D                          & B                          & -                          & -                          \\
kro124p.3 & 53915 (1832.6) & 51411 (321.2) & 53313 (1678.5) & 51351 (963.8) & B                          & -                          & D                          & B                          & -                          & D                          \\
kro124p.4 & 80373 (1120.1) & 79708 (922.3) & 81204 (1064.1) & 78973 (746.0) & -                          & -                          & D                          & B                          & -                          & D                          \\
prob.100  & 1485 (87.8)    & 1489 (75.2)   & 1505 (76.6)    & 1438 (66.8)   & -                          & -                          & -                          & -                          & D                          & D                          \\
rbg109a   & 1111 (9.9)     & 1107 (11.4)   & 1093 (9.7)     & 1067 (7.9)    & -                          & C                          & D                          & C                          & D                          & D                          \\
rbg150a   & 1872 (11.8)    & 1855 (9.7)    & 1817 (11.6)    & 1788 (10.5)   & -                          & C                          & D                          & C                          & D                          & D                          \\
rbg253a   & 3156 (15.1)    & 3123 (13.6)   & 3101 (19.8)    & 3041 (9.0)    & B                          & C                          & D                          & -                          & D                          & D                          \\
rbg341a   & 3103 (60.8)    & 2989 (40.8)   & 2990 (37.2)    & 2821 (31.5)   & B                          & C                          & D                          & -                          & D                          & D                          \\
rbg323a   & 3590 (29.9)    & 3517 (21.0)   & 3540 (27.8)    & 3393 (31.3)   & B                          & C                          & D                          & -                          & D                          & D                          \\
rbg358a   & 3284 (81.1)    & 3128 (34.9)   & 3087 (52.0)    & 2883 (36.3)   & B                          & C                          & D                          & -                          & D                          & D                          \\
rbg378a   & 3483 (54.0)    & 3347 (38.9)   & 3442 (59.8)    & 3184 (37.5)   & B                          & -                          & D                          & B                          & D                          & D                          \\ \bottomrule
\end{tabular}%
}
\end{table}

\begin{figure}[!ht]
    \centering
    \includegraphics{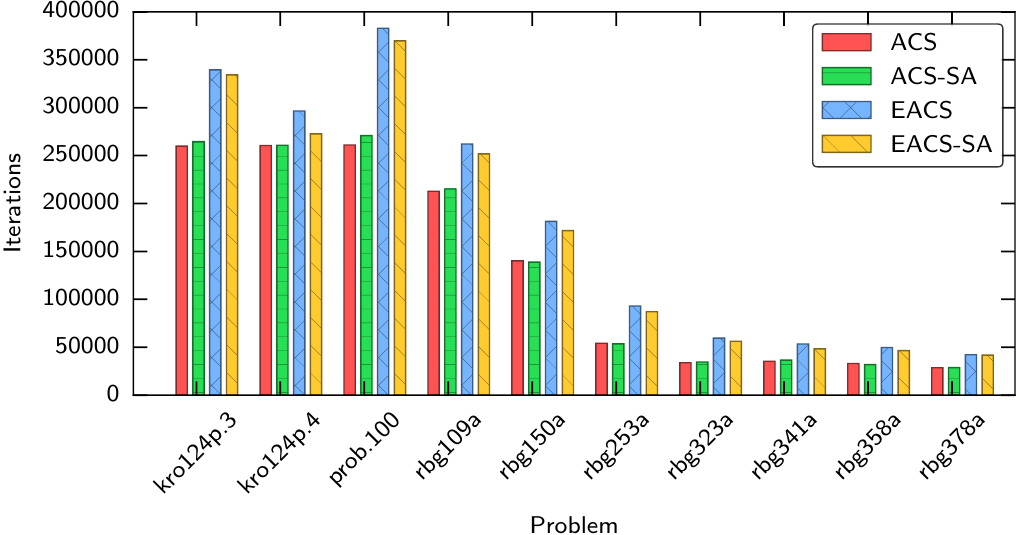}
    \caption{
    Boxplot of the mean number of iterations made by the ACS, ACS-SA, EACS and EACS-SA algorithms 
    for SOP instances selected from the TSPLIB repository.
    }
    \label{fig:iter-cmp}
\end{figure}

The Simulated Annealing component in both the ACS-SA and the EACS-SA does not increase the asymptotic time complexity of the algorithms. Only the initial temperature calculation requires a number of random solutions to be constructed, while the main ACS loop is little affected by the Metropolis rule and the cooling schedule computations. Figure~\ref{fig:iter-cmp} shows the mean number of iterations performed by each of the considered algorithms within a time limit of 60 sec. As can be observed, the number of iterations depends mostly on the size of the problem instance, while the differences between the algorithms are relatively small. The EACS and EACS-SA are faster than the other two algorithms due to the less expensive solution construction process which builds a new solution by reusing significant parts of a solution from the previous iteration.

\subsection{SOP-3-exchange-SA Parameter Tuning}
\label{sec:SOP_3_exchange_SA_Parameter_Tuning}

Similarly to the ACS-SA and EACS-SA the SOP-3-exchange-SA algorithm has two more, SA-related, parameters, namely $\lambda_{\rm LS}$ and $\gamma_{\rm LS}$.
Based on preliminary computations, several values were preselected for further investigation, namely $\lambda_{\rm LS} \in \{ 0.8, 0.9, 0.95, 0.99 \}$ and $\gamma_{\rm LS} \in \{ 0.1, 0.5, 0.9 \}$.
All 12 combinations of the parameters values were considered.
For each combination the EACS algorithm with the SOP-3-exchange-SA was run on a set of 14 instances of sizes from 400 to 700 selected from the SOPLIB2006 repository, namely: \emph{R.400.100.15, R.400.100.30, R.400.1000.15,
R.400.1000.30, R.500.100.15, R.500.1000.1, R.500.1000.15, R.500.1000.30,
R.600.100.15, R.600.100.30, R.600.1000.15, R.700.100.30, R.700.1000.1,
R.700.1000.15}.

The non-parametric Mack-Skillings test was used to verify if there were any significant differences between the results for the different $\lambda_{\rm LS}$ and $\gamma_{\rm LS}$ values, similarly to Sec.~\ref{sec:acs-sa-parameters-tuning}.
The null hypothesis $H_0$ stating that there were no differences between the medians of the solutions' quality produced for the different parameter values was rejected if $MS \ge ms_\alpha$, where $MS$ is the Mack-Skillings statistic and $ms_\alpha$ is the critical value at specified level of significance $\alpha$.
In our case, $MS \approx 1840.75$ and $ms_{0.05} \approx 19.66$, hence $H_0$ was rejected, providing strong evidence for the significant differences between the quality of results of the EACS-SA with SOP-3-exchange-SA obtained for the different $\lambda_{\rm LS}$ and $\gamma_{\rm LS}$ values.

A post-hoc multiple comparison test by Mack and Skillings~\cite{a44} was applied to find out for which values of the parameters the results were of better quality.
Table~\ref{tab:sop-3-sa-post-hoc} contains the computed $p$-values, where a value at the intersection of the $i$-th row and $j$-th column denotes the $p$-value of the comparison between the results obtained for values of $\lambda_{\rm LS}$ and $\gamma_{\rm LS}$ corresponding to the $i$-th row and $j$-th column, respectively.
An analysis of the test results revealed that for $\lambda_{\rm LS} = 0.99$ and $\gamma_{\rm LS} = 0.1$ the results were significantly better than for any other combination of values.
Simultaneously, the worst configuration, in terms of solution quality, was $\lambda_{\rm LS} = 0.8$ and $\gamma_{\rm LS} = 0.1$, hence the $\gamma_{\rm LS}$ parameter is of lower significance than $\lambda_{\rm LS}$, which directly influences the speed of the temperature decrease in the SA component of the SOP-3-exchange-SA.
Generally, the best results were obtained for $\lambda_{\rm LS}$ equal to 0.95 and 0.99.

\begin{table}[]
\centering
\caption{Table containing the p-values of the post-hoc pairwise comparison between the results of the EACS-SA with the \emph{SOP-3-exchange-SA} local search algorithm with various
$(\lambda_{\rm LS}, \gamma_{\rm LS})$ values according to the non-parametric, two-sided multiple comparison 
procedure by Mack and Skillings at $\alpha=0.05$~\cite{a44}.
The +/- symbol after a value denotes that the results for the configuration in a row were significantly better/worse than those obtained for the configuration in a column.
}
\label{tab:sop-3-sa-post-hoc}
\resizebox{\textwidth}{!}{%
\begin{tabular}{@{}cllllllllllll@{}}
\toprule
 & \multicolumn{1}{c}{\begin{tabular}[c]{@{}c@{}}A\\ (0.8, 0.1)\end{tabular}} & \multicolumn{1}{c}{\begin{tabular}[c]{@{}c@{}}B\\ (0.8, 0.5)\end{tabular}} & \multicolumn{1}{c}{\begin{tabular}[c]{@{}c@{}}C\\ (0.8, 0.9)\end{tabular}} & \multicolumn{1}{c}{\begin{tabular}[c]{@{}c@{}}D\\ (0.9, 0.1)\end{tabular}} & \multicolumn{1}{c}{\begin{tabular}[c]{@{}c@{}}E\\ (0.9, 0.5)\end{tabular}} & \multicolumn{1}{c}{\begin{tabular}[c]{@{}c@{}}F\\ (0.9, 0.9)\end{tabular}} & \multicolumn{1}{c}{\begin{tabular}[c]{@{}c@{}}G\\ (0.95,0.1)\end{tabular}} & \multicolumn{1}{c}{\begin{tabular}[c]{@{}c@{}}H\\ (0.95,0.5)\end{tabular}} & \multicolumn{1}{c}{\begin{tabular}[c]{@{}c@{}}I\\ (0.95, 0.9)\end{tabular}} & \multicolumn{1}{c}{\begin{tabular}[c]{@{}c@{}}J\\ (0.99, 0.1)\end{tabular}} & \multicolumn{1}{c}{\begin{tabular}[c]{@{}c@{}}K\\ (0.99, 0.5)\end{tabular}} & \multicolumn{1}{c}{\begin{tabular}[c]{@{}c@{}}L\\ (0.99, 0.9)\end{tabular}} \\ \midrule
A & -- & $<0.0001$- & $<0.0001$- & 0.0131- & $<0.0001$- & $<0.0001$- & $<0.0001$- & $<0.0001$- & $<0.0001$- & $<0.0001$- & $<0.0001$- & $<0.0001$- \\
B & $<0.0001$+ & -- & $<0.0001$- & 0.8878 & 0.0001- & $<0.0001$- & $<0.0001$- & $<0.0001$- & $<0.0001$- & $<0.0001$- & $<0.0001$- & $<0.0001$- \\
C & $<0.0001$+ & $<0.0001$+ & -- & $<0.0001$+ & 1.0000 & $<0.0001$- & 1.0000 & $<0.0001$- & $<0.0001$- & $<0.0001$- & $<0.0001$- & $<0.0001$- \\
D & 0.0131+ & 0.8878 & $<0.0001$- & -- & $<0.0001$- & $<0.0001$- & $<0.0001$- & $<0.0001$- & $<0.0001$- & $<0.0001$- & $<0.0001$- & $<0.0001$- \\
E & $<0.0001$+ & 0.0001+ & 1.0000 & $<0.0001$+ & -- & $<0.0001$- & 0.9990 & $<0.0001$- & $<0.0001$- & $<0.0001$- & $<0.0001$- & $<0.0001$- \\
F & $<0.0001$+ & $<0.0001$+ & $<0.0001$+ & $<0.0001$+ & $<0.0001$+ & -- & $<0.0001$+ & 0.0309- & $<0.0001$- & $<0.0001$- & $<0.0001$- & $<0.0001$- \\
G & $<0.0001$+ & $<0.0001$+ & 1.0000 & $<0.0001$+ & 0.9990 & $<0.0001$- & -- & $<0.0001$- & $<0.0001$- & $<0.0001$- & $<0.0001$- & $<0.0001$- \\
H & $<0.0001$+ & $<0.0001$+ & $<0.0001$+ & $<0.0001$+ & $<0.0001$+ & 0.0309+ & $<0.0001$+ & -- & 0.0077- & $<0.0001$- & 0.0014- & 0.3709 \\
I & $<0.0001$+ & $<0.0001$+ & $<0.0001$+ & $<0.0001$+ & $<0.0001$+ & $<0.0001$+ & $<0.0001$+ & 0.0077+ & -- & 0.0073- & 1.0000 & 0.9712 \\
J & $<0.0001$+ & $<0.0001$+ & $<0.0001$+ & $<0.0001$+ & $<0.0001$+ & $<0.0001$+ & $<0.0001$+ & $<0.0001$+ & 0.0073+ & -- & 0.0331+ & $<0.0001$+ \\
K & $<0.0001$+ & $<0.0001$+ & $<0.0001$+ & $<0.0001$+ & $<0.0001$+ & $<0.0001$+ & $<0.0001$+ & 0.0014+ & 1.0000 & 0.0331- & -- & 0.8270 \\
L & $<0.0001$+ & $<0.0001$+ & $<0.0001$+ & $<0.0001$+ & $<0.0001$+ & $<0.0001$+ & $<0.0001$+ & 0.3709 & 0.9712 & $<0.0001$- & 0.8270 & -- \\ \bottomrule
\end{tabular}%
}
\end{table}

\subsection{Comparison of algorithms}
\label{sec:Comparison_of_algorithms}

The last part of the experiments concerned the performance of ACS, ACS-SA, EACS and EACS-SA combined with the LS algorithms, i.e. SOP-3-exchange and SOP-3-exchange-SA. This gives a total of 8 algorithm combinations.
To make the comparison fair, the algorithms were run with the same time limit of 120 seconds and the same values of parameters (where appropriate). The algorithms were run on SOP instances (48 in total) from the SOPLIB2006 repository~\cite{a23}.
It is worth noting, that the EACS with the SOP-3-exchange is the current state-of-the-art metaheuristic for the SOP~\cite{a12}.

\begin{table}[]
\centering
\scriptsize
\caption{Table containing the p-values of the post-hoc pairwise comparison between the results of the 
ACS, ACS-SA, EACS and EACS-SA algorithms according to the non-parametric, two-sided multiple comparison 
procedure by Mack and Skillings at $\alpha=0.05$~\cite{a44}.
The +/- symbol after a value denotes that the results for the algorithm in a row were significantly better/worse than those obtained for the algorithm in a column.
The \emph{LS1} and \emph{LS2} subscripts denote the local search method used, i.e. the \emph{SOP-3-exchange} and \emph{SOP-3-exchange-SA}, respectively.
}
\label{tab:alg-to-alg-p-values}
\begin{tabular}{@{}lllllllll@{}}
\toprule
Algorithm           & \multicolumn{1}{c}{A} & \multicolumn{1}{c}{B} & \multicolumn{1}{c}{C} & \multicolumn{1}{c}{D} & \multicolumn{1}{c}{E} & \multicolumn{1}{c}{F} & \multicolumn{1}{c}{G} & \multicolumn{1}{c}{H} \\ \midrule
ACS$_{\rm LS1}$ (A)          & -                     & \textless0.0001-      & \textless0.0001+      & \textless0.0001+      & \textless0.0001-      & \textless0.0001-      & \textless0.0001-      & \textless0.0001-      \\
ACS$_{\rm LS2}$ (B)     & \textless0.0001+      & -                     & \textless0.0001+      & \textless0.0001+      & \textless0.0001-      & \textless0.0001-      & \textless0.0001-      & \textless0.0001-      \\
ACS-SA$_{\rm LS1}$ (C)       & \textless0.0001-      & \textless0.0001-      & -                     & 1.0000                & \textless0.0001-      & \textless0.0001-      & \textless0.0001-      & \textless0.0001-      \\
ACS-SA$_{\rm LS2}$ (D)  & \textless0.0001-      & \textless0.0001-      & 1.0000                & -                     & \textless0.0001-      & \textless0.0001-      & \textless0.0001-      & \textless0.0001-      \\
EACS$_{\rm LS1}$ (E)     & \textless0.0001+      & \textless0.0001+      & \textless0.0001+      & \textless0.0001+      & -                     & \textless0.0001-      & 0.9999                & \textless0.0001-      \\
EACS$_{\rm LS2}$ (F)    & \textless0.0001+      & \textless0.0001+      & \textless0.0001+      & \textless0.0001+      & \textless0.0001+      & -                     & \textless0.0001+      & \textless0.0001+      \\
EACS-SA$_{\rm LS1}$ (G)  & \textless0.0001+      & \textless0.0001+      & \textless0.0001+      & \textless0.0001+      & 0.9999                & \textless0.0001-      & -                     & \textless0.0001-      \\
EACS-SA$_{\rm LS2}$ (H) & \textless0.0001+      & \textless0.0001+      & \textless0.0001+      & \textless0.0001+      & \textless0.0001+      & \textless0.0001-      & \textless0.0001+      & -                     \\ \bottomrule
\end{tabular}
\end{table}

A quick analysis of the obtained results showed noticeable differences in the efficiency of the investigated algorithms.
The experiment design allows to check for statistically significant differences between the algorithms by using the non-parametric Mack-Skillings test for a two-factor layout. The first factor is the algorithm that is applied while the second (blocking) factor is the SOP instance that is solved.
The null hypothesis $H_0$ of our interest is that there are no differences in the quality of the solutions generated by the algorithms.
The rejection of $H_0$ would mean that the algorithms differ in the quality of generated solutions. The critical value for the test at level of significance equal to 0.05 is $ms_{0.05} \approx 14.03$ and the Mack-Skillings statistic is $MS \approx 8843.11$, meaning that $MS > ms_{0.05}$, hence $H_0$ was rejected.

Rejection of the null hypothesis allows us to apply a post-hoc test (also proposed by Mack and Skillings~\cite{a44}) to perform a pairwise comparison of the algorithms. 
The resulting $p$-values are shown in Tab.~\ref{tab:alg-to-alg-p-values}.
As can be observed, all values are either close to 0 or close to 1, meaning that the differences are either sharp or nonexistent, respectively.
Not surprisingly, all EACS variants obtained significantly better results than the ACS-based algorithms. The most efficient algorithm was the EACS with the SOP-3-exchange-SA LS, which obtained results that were significantly better than any of the other remaining algorithms. The second best was the EACS-SA with the SOP-3-exchange-SA LS.
Out of the four ACS variants the ACS with SOP-3-exchange-SA performed better than the other three, thus confirming the efficiency of the proposed SOP-3-exchange-SA LS.
The worst performing were the ACS-SA with the SOP-3-exchange and the ACS-SA with the SOP-3-exchange-SA.
The poor performance of the ACS-SA variants can be explained by the weakened emphasis on the exploitation which admittedly increases the probability of escaping from local optima but also slows the overall convergence of the algorithm, which is clearly visible if the computational budget is modest, as in the experiment conducted here (120 sec.).

\begin{table}[]
\scriptsize
\centering
\caption{Sample mean and standard deviation values for the EACS and EACS-SA algorithms 
with the SOP-3-exchange and SOP-3-exchange-SA LS heuristics obtained for the SOP instances from the SOPLIB2006 repository~\cite{a23}. The smallest means in a row are written in bold.}
\label{tab:eacs-cmp-avg}
\begin{tabular}{@{}lrrrrrrrr@{}}
\toprule
\multicolumn{1}{c}{\multirow{2}{*}{Instance}} & \multicolumn{2}{c}{\begin{tabular}[c]{@{}c@{}}EACS \\+SOP-3-exchange\end{tabular}} & \multicolumn{2}{c}{\begin{tabular}[c]{@{}c@{}}EACS \\+SOP-3-exchange-SA\end{tabular}} & \multicolumn{2}{c}{\begin{tabular}[c]{@{}c@{}}EACS-SA \\+SOP-3-exchange\end{tabular}} & \multicolumn{2}{c}{\begin{tabular}[c]{@{}c@{}}EACS-SA \\+SOP-3-exchange-SA\end{tabular}} \\
\multicolumn{1}{c}{}                          & \multicolumn{1}{c}{Avg.}               & \multicolumn{1}{c}{Std. dev.}               & \multicolumn{1}{c}{Avg.}                 & \multicolumn{1}{c}{Std. dev.}                & \multicolumn{1}{c}{Avg.}                 & \multicolumn{1}{c}{Std. dev.}                & \multicolumn{1}{c}{Avg.}                  & \multicolumn{1}{c}{Std. dev.}                  \\ \midrule
R.200.100.1                                   & 74.5                                   & 3.8                                         & 69.4                                     & 2.4                                          & 67.6                                     & 1.9                                          & \textbf{66.6}                             & 2.2                                            \\
R.200.100.15                                  & 1935.5                                 & 31.0                                        & 1837.1                                   & 25.0                                         & 1916.9                                   & 27.8                                         & \textbf{1828.6}                           & 56.5                                           \\
R.200.100.30                                  & \textbf{4216.0}                        & 0.0                                         & \textbf{4216.0}                          & 0.0                                          & \textbf{4216.0}                          & 0.0                                          & \textbf{4216.0}                           & 0.0                                            \\
R.200.100.60                                  & \textbf{71749.0}                       & 0.0                                         & \textbf{71749.0}                         & 0.0                                          & \textbf{71749.0}                         & 0.0                                          & \textbf{71749.0}                          & 0.0                                            \\
R.200.1000.1                                  & 1457.2                                 & 17.9                                        & 1459.0                                   & 21.5                                         & \textbf{1433.9}                          & 9.8                                          & 1467.2                                    & 12.2                                           \\
R.200.1000.15                                 & 21766.2                                & 298.1                                       & \textbf{20771.4}                         & 201.0                                        & 21577.6                                  & 399.5                                        & 21104.0                                   & 881.0                                          \\
R.200.1000.30                                 & \textbf{41196.0}                       & 0.0                                         & \textbf{41196.0}                         & 0.0                                          & \textbf{41196.0}                         & 0.0                                          & \textbf{41196.0}                          & 0.0                                            \\
R.200.1000.60                                 & \textbf{71556.0}                       & 0.0                                         & \textbf{71556.0}                         & 0.0                                          & \textbf{71556.0}                         & 0.0                                          & \textbf{71556.0}                          & 0.0                                            \\
\midrule
R.300.100.1                                   & 45.2                                   & 3.9                                         & 36.0                                     & 2.7                                          & 39.7                                     & 3.2                                          & \textbf{31.3}                             & 2.4                                            \\
R.300.100.15                                  & 3328.1                                 & 54.8                                        & \textbf{3177.4}                          & 18.7                                         & 3289.9                                   & 31.0                                         & 3209.9                                    & 97.4                                           \\
R.300.100.30                                  & 6124.2                                 & 10.9                                        & \textbf{6120.0}                          & 0.0                                          & 6122.3                                   & 5.3                                          & \textbf{6120.0}                           & 0.0                                            \\
R.300.100.60                                  & \textbf{9726.0}                        & 0.0                                         & \textbf{9726.0}                          & 0.0                                          & \textbf{9726.0}                          & 0.0                                          & \textbf{9726.0}                           & 0.0                                            \\
R.300.1000.1                                  & \textbf{1424.0}                        & 29.4                                        & 1432.2                                   & 33.3                                         & 1436.7                                   & 21.6                                         & 1464.2                                    & 24.3                                           \\
R.300.1000.15                                 & 31556.9                                & 647.4                                       & \textbf{29713.0}                         & 441.0                                        & 31196.1                                  & 637.0                                        & 30002.0                                   & 1319.5                                         \\
R.300.1000.30                                 & 54223.3                                & 72.7                                        & \textbf{54147.0}                         & 0.0                                          & 54176.5                                  & 30.7                                         & \textbf{54147.0}                          & 0.0                                            \\
R.300.1000.60                                 & 109482.9                               & 36.3                                        & \textbf{109471.0}                        & 0.0                                          & 109488.3                                 & 41.3                                         & \textbf{109471.0}                         & 0.0                                            \\
\midrule
R.400.100.1                                   & 35.9                                   & 4.8                                         & 23.6                                     & 4.1                                          & 33.1                                     & 4.6                                          & \textbf{19.0}                             & 2.2                                            \\
R.400.100.15                                  & 4270.9                                 & 67.9                                        & 3969.8                                   & 29.3                                         & 4184.8                                   & 52.5                                         & \textbf{3930.5}                           & 28.2                                           \\
R.400.100.30                                  & 8167.6                                 & 9.9                                         & \textbf{8165.0}                          & 0.0                                          & 8166.0                                   & 0.2                                          & \textbf{8165.0}                           & 0.0                                            \\
R.400.100.60                                  & \textbf{15228.0}                       & 0.0                                         & \textbf{15228.0}                         & 0.0                                          & \textbf{15228.0}                         & 0.0                                          & \textbf{15228.0}                          & 0.0                                            \\
R.400.1000.1                                  & \textbf{1504.7}                        & 29.9                                        & 1528.6                                   & 24.8                                         & 1521.9                                   & 22.6                                         & 1561.3                                    & 31.9                                           \\
R.400.1000.15                                 & 42436.7                                & 632.7                                       & 39854.6                                  & 317.6                                        & 41258.3                                  & 516.9                                        & \textbf{39293.0}                          & 204.3                                          \\
R.400.1000.30                                 & 85320.8                                & 117.5                                       & 85157.5                                  & 62.9                                         & 85188.6                                  & 77.1                                         & \textbf{85128.6}                          & 3.3                                            \\
R.400.1000.60                                 & \textbf{140816.0}                      & 0.0                                         & \textbf{140816.0}                        & 0.0                                          & \textbf{140816.0}                        & 0.0                                          & \textbf{140816.0}                         & 0.0                                            \\
\midrule
R.500.100.1                                   & 25.4                                   & 3.3                                         & 11.6                                     & 2.6                                          & 25.3                                     & 4.4                                          & \textbf{9.9}                              & 2.6                                            \\
R.500.100.15                                  & 5801.2                                 & 91.3                                        & \textbf{5417.1}                          & 54.5                                         & 5717.8                                   & 86.4                                         & 5449.8                                    & 254.7                                          \\
R.500.100.30                                  & 9709.9                                 & 20.4                                        & \textbf{9668.3}                          & 5.4                                          & 9696.0                                   & 13.2                                         & 9740.5                                    & 87.2                                           \\
R.500.100.60                                  & 18255.3                                & 5.2                                         & \textbf{18240.0}                         & 0.0                                          & 18253.4                                  & 7.7                                          & \textbf{18240.0}                          & 0.0                                            \\
R.500.1000.1                                  & 1524.8                                 & 38.8                                        & 1556.9                                   & 34.3                                         & \textbf{1542.5}                          & 34.1                                         & 1591.3                                    & 38.3                                           \\
R.500.1000.15                                 & 54644.0                                & 760.2                                       & 50503.3                                  & 360.3                                        & 53719.0                                  & 632.9                                        & \textbf{50310.4}                          & 420.0                                          \\
R.500.1000.30                                 & 99465.1                                & 209.0                                       & 99038.0                                  & 42.4                                         & 99166.6                                  & 78.9                                         & \textbf{99022.0}                          & 160.9                                          \\
R.500.1000.60                                 & 178247.6                               & 108.6                                       & \textbf{178212.0}                        & 0.0                                          & 178268.9                                 & 130.1                                        & \textbf{178212.0}                         & 0.0                                            \\
\midrule
R.600.100.1                                   & 21.3                                   & 4.0                                         & \textbf{4.9}                             & 2.1                                          & 22.6                                     & 2.9                                          & 14.3                                      & 8.8                                            \\
R.600.100.15                                  & 6280.1                                 & 114.5                                       & \textbf{5621.5}                          & 40.9                                         & 6224.2                                   & 100.5                                        & 5797.5                                    & 511.5                                          \\
R.600.100.30                                  & 12504.9                                & 15.7                                        & \textbf{12467.9}                         & 5.2                                          & 12499.7                                  & 11.8                                         & 12569.7                                   & 34.9                                           \\
R.600.100.60                                  & \textbf{23293.0}                       & 0.0                                         & \textbf{23293.0}                         & 0.0                                          & \textbf{23293.0}                         & 0.0                                          & \textbf{23293.0}                          & 0.0                                            \\
R.600.1000.1                                  & \textbf{1582.7}                        & 36.0                                        & 1679.3                                   & 40.6                                         & 1657.7                                   & 104.0                                        & 1749.3                                    & 47.2                                           \\
R.600.1000.15                                 & 61492.4                                & 964.8                                       & \textbf{56638.3}                         & 389.9                                        & 62859.3                                  & 3426.5                                       & 61581.3                                   & 7760.0                                         \\
R.600.1000.30                                 & 127460.6                               & 384.8                                       & \textbf{126798.6}                        & 3.5                                          & 128011.6                                 & 305.2                                        & 129319.9                                  & 505.6                                          \\
R.600.1000.60                                 & \textbf{214608.0}                      & 0.0                                         & \textbf{214608.0}                        & 0.0                                          & 214641.1                                 & 59.0                                         & \textbf{214608.0}                         & 0.0                                            \\
\midrule
R.700.100.1                                   & 16.6                                   & 3.5                                         & \textbf{1.7}                             & 0.7                                          & 26.4                                     & 6.0                                          & 30.3                                      & 7.5                                            \\
R.700.100.15                                  & 7835.3                                 & 110.4                                       & \textbf{7195.4}                          & 65.3                                         & 8964.0                                   & 775.0                                        & 7865.6                                    & 854.7                                          \\
R.700.100.30                                  & 14585.7                                & 28.2                                        & \textbf{14510.3}                         & 0.8                                          & 14693.2                                  & 36.5                                         & 14792.3                                   & 37.7                                           \\
R.700.100.60                                  & 24108.7                                & 7.3                                         & \textbf{24102.0}                         & 0.0                                          & 24131.3                                  & 13.6                                         & \textbf{24102.0}                          & 0.0                                            \\
R.700.1000.1                                  & \textbf{1539.8}                        & 45.9                                        & 1644.1                                   & 36.0                                         & 1718.5                                   & 129.8                                        & 1930.4                                    & 83.8                                           \\
R.700.1000.15                                 & 72271.5                                & 934.4                                       & \textbf{66738.6}                         & 479.7                                        & 79815.4                                  & 7934.4                                       & 83061.5                                   & 9842.3                                         \\
R.700.1000.30                                 & 135922.5                               & 441.5                                       & \textbf{134495.7}                        & 40.0                                         & 136468.1                                 & 347.6                                        & 138703.9                                  & 477.9                                          \\
R.700.1000.60                                 & 245602.3                               & 49.5                                        & \textbf{245589.0}                        & 0.0                                          & 245848.2                                 & 154.1                                        & \textbf{245589.0}                         & 0.0                                            \\ \bottomrule
\end{tabular}
\end{table}

Even though some of the algorithms can be seen as generally more efficient than others, this is not true in every case,
as can be observed in Tab.~\ref{tab:eacs-cmp-avg}, in which the sample mean and sample standard deviation values are presented for the EACS and EACS-SA algorithms.
The two most efficient, in terms of solution quality, were the EACS with the SOP-3-exchange and the EACS-SA with the SOP-3-exchange LS. While the former achieved lower mean values for more problem instances, the latter performed particularly well for instances of a size up to 500. For the largest instances (\emph{R.600.*} and \emph{R.700.*}), the EACS with the SOP-3-exchange obtained the lowest mean values in 14 out of 16 cases. This suggests that the EACS-SA algorithm did not have enough time to converge within the specified time limit.

Similar observations can be made from the analysis of the best solutions found by the algorithms presented in Tab.~\ref{tab:eacs-cmp-best}.
The table also contains the values of the best-known solutions; some of which were obtained by Gouveia and Ruthmair by using an exact method (branch-and-cut)~\cite{a13} and by Papapanagiotou et al.~\cite{papapanagiotou2015comparison}, while the rest by metaheuristics, including the EACS with the SOP-3-exchange~\cite{a12}.
For the 18 SOP instances, all four algorithms were able to find the best-known solution at least once per 30 runs.
For the 10 SOP instances new best solutions were found by the proposed algorithms. The EACS with the SOP-3-exchange-SA found the new best solutions for 6 instances, i.e. \emph{R.300.1000.15, R.500.100.15, R.500.100.15,
R.600.1000.15, R.700.100.15}, and \emph{R.700.1000.15}. The EACS-SA with the SOP-3-exchange-SA found the new best solutions for 4 instances, namely:
\emph{R.300.100.15, R.400.100.15, R.400.1000.15} and \emph{R.500.1000.15}.
Overall, the best known or improved solutions were obtained by at least one of the algorithms in 37 out of 48 cases (77\%).
All algorithms struggled most with instances in which the number of precedence constraints was smallest, i.e. 1\% (instances \emph{R.*.*.1}) which suggests that there is still some room for improvement of the LS algorithms.

\begin{table}[]
\scriptsize
\centering
\caption{Best solution values obtained by the EACS and EACS-SA algorithms 
with the SOP-3-exchange and SOP-3-exchange-SA LS heuristics for the SOP instances from the SOPLIB2006 repository~\cite{a23}. The smallest values in a row were written in bold.
Values improving upon the best results known from the literature were underlined.
}
\label{tab:eacs-cmp-best}
\begin{tabular}{@{}lrrrrr@{}}
\toprule
\multicolumn{1}{c}{Instance} & \multicolumn{1}{c}{\begin{tabular}[c]{@{}c@{}}Best known\\\cite{a12,a13,papapanagiotou2015comparison}\end{tabular}} & \multicolumn{1}{c}{\begin{tabular}[c]{@{}c@{}}EACS\\+SOP-3-exchange\end{tabular}} & \multicolumn{1}{c}{\begin{tabular}[c]{@{}c@{}}EACS\\+SOP-3-exchange-SA\end{tabular}} & \multicolumn{1}{c}{\begin{tabular}[c]{@{}c@{}}EACS-SA\\+SOP-3-exchange\end{tabular}} & \multicolumn{1}{c}{\begin{tabular}[c]{@{}c@{}}EACS-SA\\+SOP-3-exchange-SA\end{tabular}} \\ \midrule
R.200.100.1                  & \textbf{61}                    & 67                                                                                  & 64                                                                                     & 63                                                                                     & 64                                                                                        \\
R.200.100.15                 & \textbf{1792}                  & 1890                                                                                & 1796                                                                                   & 1868                                                                                   & \textbf{1792}                                                                             \\
R.200.100.30                 & \textbf{4216}                  & \textbf{4216}                                                                       & \textbf{4216}                                                                          & \textbf{4216}                                                                          & \textbf{4216}                                                                             \\
R.200.100.60                 & \textbf{71749}                 & \textbf{71749}                                                                      & \textbf{71749}                                                                         & \textbf{71749}                                                                         & \textbf{71749}                                                                            \\
R.200.1000.1                 & \textbf{1404}                  & 1426                                                                                & 1423                                                                                   & 1416                                                                                   & 1437                                                                                      \\
R.200.1000.15                & \textbf{20481}                 & 21113                                                                               & \textbf{20481}                                                                         & 20946                                                                                  & \textbf{20481}                                                                            \\
R.200.1000.30                & \textbf{41196}                 & \textbf{41196}                                                                      & \textbf{41196}                                                                         & \textbf{41196}                                                                         & \textbf{41196}                                                                            \\
R.200.1000.60                & \textbf{71556}                 & \textbf{71556}                                                                      & \textbf{71556}                                                                         & \textbf{71556}                                                                         & \textbf{71556}                                                                            \\
\midrule
R.300.100.1                  & \textbf{26}                    & 39                                                                                  & 31                                                                                     & 32                                                                                     & 28                                                                                        \\
R.300.100.15                 & 3161                           & 3251                                                                                & 3154                                                                                   & 3235                                                                                   & {\ul \textbf{3152}}                                                                       \\
R.300.100.30                 & \textbf{6120}                  & \textbf{6120}                                                                       & \textbf{6120}                                                                          & \textbf{6120}                                                                          & \textbf{6120}                                                                             \\
R.300.100.60                 & \textbf{9726}                  & \textbf{9726}                                                                       & \textbf{9726}                                                                          & \textbf{9726}                                                                          & \textbf{9726}                                                                             \\
R.300.1000.1                 & \textbf{1294}                  & 1363                                                                                & 1382                                                                                   & 1389                                                                                   & 1413                                                                                      \\
R.300.1000.15                & 29183                          & 30295                                                                               & {\ul \textbf{29068}}                                                                   & 30099                                                                                  & 29111                                                                                     \\
R.300.1000.30                & \textbf{54147}                 & \textbf{54147}                                                                      & \textbf{54147}                                                                         & \textbf{54147}                                                                         & \textbf{54147}                                                                            \\
R.300.1000.60                & \textbf{109471}                & \textbf{109471}                                                                     & \textbf{109471}                                                                        & \textbf{109471}                                                                        & \textbf{109471}                                                                           \\
\midrule
R.400.100.1                  & \textbf{13}                    & 28                                                                                  & 17                                                                                     & 26                                                                                     & 14                                                                                        \\
R.400.100.15                 & 3906                           & 4139                                                                                & 3918                                                                                   & 4088                                                                                   & {\ul \textbf{3883}}                                                                       \\
R.400.100.30                 & \textbf{8165}                  & \textbf{8165}                                                                       & \textbf{8165}                                                                          & \textbf{8165}                                                                          & \textbf{8165}                                                                             \\
R.400.100.60                 & \textbf{15228}                 & \textbf{15228}                                                                      & \textbf{15228}                                                                         & \textbf{15228}                                                                         & \textbf{15228}                                                                            \\
R.400.1000.1                 & \textbf{1343}                  & 1459                                                                                & 1488                                                                                   & 1491                                                                                   & 1496                                                                                      \\
R.400.1000.15                & 43268                          & 41067                                                                               & 39148                                                                                  & 40457                                                                                  & {\ul \textbf{38963}}                                                                      \\
R.400.1000.30                & \textbf{85128}                 & \textbf{85128}                                                                      & \textbf{85128}                                                                         & \textbf{85128}                                                                         & \textbf{85128}                                                                            \\
R.400.1000.60                & \textbf{140816}                & \textbf{140816}                                                                     & \textbf{140816}                                                                        & \textbf{140816}                                                                        & \textbf{140816}                                                                           \\
\midrule
R.500.100.1                  & \textbf{4}                     & 17                                                                                  & 8                                                                                      & 17                                                                                     & 6                                                                                         \\
R.500.100.15                 & 5361                           & 5615                                                                                & {\ul \textbf{5305}}                                                                    & 5575                                                                                   & 5315                                                                                      \\
R.500.100.30                 & \textbf{9665}                  & 9687                                                                                & \textbf{9665}                                                                          & 9675                                                                                   & \textbf{9665}                                                                             \\
R.500.100.60                 & \textbf{18240}                 & \textbf{18240}                                                                      & \textbf{18240}                                                                         & \textbf{18240}                                                                         & \textbf{18240}                                                                            \\
R.500.1000.1                 & \textbf{1316}                  & 1464                                                                                & 1487                                                                                   & 1476                                                                                   & 1519                                                                                      \\
R.500.1000.15                & 50725                          & 53082                                                                               & 49907                                                                                  & 52630                                                                                  & {\ul \textbf{49719}}                                                                      \\
R.500.1000.30                & \textbf{98987}                 & 99072                                                                               & \textbf{98987}                                                                         & 99002                                                                                  & \textbf{98987}                                                                            \\
R.500.1000.60                & \textbf{178212}                & \textbf{178212}                                                                     & \textbf{178212}                                                                        & \textbf{178212}                                                                        & \textbf{178212}                                                                           \\
\midrule
R.600.100.1                  & \textbf{1}                     & 15                                                                                  & 2                                                                                      & 16                                                                                     & 4                                                                                         \\
R.600.100.15                 & 5684                           & 6087                                                                                & {\ul \textbf{5548}}                                                                    & 6008                                                                                   & 5568                                                                                      \\
R.600.100.30                 & \textbf{12465}                 & 12481                                                                               & \textbf{12465}                                                                         & 12484                                                                                  & 12496                                                                                     \\
R.600.100.60                 & \textbf{23293}                 & \textbf{23293}                                                                      & \textbf{23293}                                                                         & \textbf{23293}                                                                         & \textbf{23293}                                                                            \\
R.600.1000.1                 & \textbf{1337}                  & 1500                                                                                & 1611                                                                                   & 1518                                                                                   & 1644                                                                                      \\
R.600.1000.15                & 57237                          & 59660                                                                               & {\ul \textbf{55725}}                                                                   & 60194                                                                                  & 56283                                                                                     \\
R.600.1000.30                & \textbf{126798}                & \textbf{126798}                                                                     & \textbf{126798}                                                                        & 127531                                                                                 & 128214                                                                                    \\
R.600.1000.60                & \textbf{214608}                & \textbf{214608}                                                                     & \textbf{214608}                                                                        & \textbf{214608}                                                                        & \textbf{214608}                                                                           \\
\midrule
R.700.100.1                  & \textbf{1}                     & 10                                                                                  & \textbf{1}                                                                             & 18                                                                                     & 16                                                                                        \\
R.700.100.15                 & 7311                           & 7629                                                                                & {\ul \textbf{7109}}                                                                    & 8011                                                                                   & 7191                                                                                      \\
R.700.100.30                 & \textbf{14510}                 & 14534                                                                               & \textbf{14510}                                                                         & 14620                                                                                  & 14714                                                                                     \\
R.700.100.60                 & \textbf{24102}                 & \textbf{24102}                                                                      & \textbf{24102}                                                                         & 24114                                                                                  & \textbf{24102}                                                                            \\
R.700.1000.1                 & \textbf{1231}                  & 1462                                                                                & 1583                                                                                   & 1588                                                                                   & 1713                                                                                      \\
R.700.1000.15                & 66837                          & 70219                                                                               & {\ul \textbf{66008}}                                                                   & 71484                                                                                  & 68009                                                                                     \\
R.700.1000.30                & \textbf{134474}                & \textbf{134712}                                                                     & \textbf{134474}                                                                        & \textbf{135841}                                                                        & \textbf{137784}                                                                           \\
R.700.1000.60                & \textbf{245589}                & \textbf{245589}                                                                     & \textbf{245589}                                                                        & \textbf{245589}                                                                        & \textbf{245589}                                                                           \\ \bottomrule
\end{tabular}
\end{table}

{\R
The results as presented above confirm that the proposed incorporation of the SA into the main algorithm (EACS) and into the local search (SOP-3-exchange) is able to improve the quality of the generated solutions to the SOP.
In order to further clarify the differences between the existing approach, i.e. the EACS with the SOP-3-exchange LS, and the proposed EACS-SA with the SOP-3-exchange-SA, both were run on SOP instances from the SOPLIB2006 repository, however the computation time was increased to 600 seconds. This is a five-fold increase vs the time limit used in the experiments presented above. 
By giving the algorithms more time, we lower the risk of one algorithm dominating an other because of the limited time.
The results are presented in Tab.~\ref{tab:cmp-ext}. 
In most cases the results of the EACS-SA with the SOP-3-exchange-SA were of a better quality than those obtained for the EACS with the SOP-3-exchange, although the relative differences between the algorithms depended on the SOP instance that was being solved.
The results were checked for a statistically significant differences using the non-parametric Wilcoxon rank sum test at a significance level of $\alpha=0.05$ (the respective $p$-values are reported in the table). 
In 33 out of 48 (69\%) cases (instances) the solutions generated by the EACS-SA with the SOP-3-exchange-SA
were of a significantly better quality than those generated by the EACS with the SOP-3-exchange.
In 4 cases (8\%) the results of the former algorithm were significantly worse and in 11 cases (23\%) no significant differences were observed.

Taking into account the best solutions generated during 30 executions of the algorithms for each of the SOP instances considered, the proposed algorithm reached the best-known results in 31 cases, and in 10 cases new best solutions were found. Because of the increased computation time limit, in 7 out of those 10 cases the results were improvement over those presented in Tab.~\ref{tab:eacs-cmp-best}. To summarize, the best-known or improved results were generated for 41 out of the 48 (85\%) SOP instances considered here.
The EACS with the SOP-3-exchange found the best known results in 18 cases; however, no new best solutions were found in any case.

All of the SOP instances for which the EACS-SA with the SOP-3-exchange-SA generated significantly worse results than the EACS with the SOP-3-exchange are of the form \emph{R.*.1000.1} what suggests either an overall inferior convergence of the former algorithm for this kind of instances, or an insufficient time limit to match the convergence of the latter algorithm.
In fact, the second possibility seems to be true because for the smallest of the \emph{R.*.1000.1} instances, i.e. \emph{R.200.1000.1}, the EACS-SA with the SOP-3-exchange-SA  generated significantly better results, and for the second smallest instance, i.e. \emph{R.300.1000.1}, there were no significant differences between the results of the two algorithms.
To confirm our assumption, both algorithms were run for the instances: \emph{R.300.1000.1, R.400.1000.1, R.500.1000.1, R.600.1000.1}, and \emph{R.700.1000.1} but with the time limit increased to 1200 seconds (doubled) per run.
The results are presented in Tab.~\ref{tab:cmp-ext-1200}.
As can be seen, the advantage of the EACS with the SOP-3-exchange over the EACS-SA with the SOP-3-exchange-SA disappeared, and both algorithms generated results of a similar quality. Statistical comparison based on the non-parametric Wilcoxon rank sum test showed no significant differences for the \emph{R.400.1000.1, R.500.1000.1} and \emph{R.600.1000.1} instances. Surprisingly, the increased time limit allowed the EACS-SA to obtain significantly better results for the two remaining instances, i.e. \emph{R.300.1000.1}, and \emph{R.700.1000.1}, although the advantage was small relative to the optimum values.

Considering all the results, the efficiency of the proposed algorithm (in terms of the quality of solutions)
was statistically significantly better than the original approach for approx. 73\% of the SOP instances, while never being worse. However, a sufficient computation time is necessary to reach this level of performance. In most cases 600 seconds was enough, whereas for a few instances the limit of 1200 seconds was necessary. 
In practical applications, the algorithm could be sped up by using parallel computations.
}

\begin{table}[]
\centering
\caption{\R Results of the EACS with the SOP-3-exchange (I) and EACS-SA with the SOP-3-exchange-SA (II) algorithms for SOPLIB2006 instances with the time limit set to 600 sec.
\emph{Verdict} denotes the algorithm for which the obtained results were of a significantly better quality than the results of the other algorithm according to the non-parametric Wilcoxon rank-sum test at a level of significance $\alpha=0.05$. Cases for which there was no significant difference are marked with a "--".
}
\label{tab:cmp-ext}
\resizebox{\textwidth}{!}{%
\begin{tabular}{@{}lrrrrrrrrc@{}}
\toprule
\multicolumn{1}{c}{\multirow{2}{*}{Instance}} & \multicolumn{1}{c}{\multirow{2}{*}{\begin{tabular}[c]{@{}c@{}}Best known\\\cite{a12,a13,papapanagiotou2015comparison}\end{tabular}}} & \multicolumn{3}{c}{EACS + SOP-3-exchange (I)}                                       & \multicolumn{3}{c}{EACS-SA + SOP-3-exchange-SA (II)}                                & \multicolumn{1}{c}{\multirow{2}{*}{$p$-value}} & \multirow{2}{*}{\begin{tabular}[c]{@{}c@{}}Verdict\\ I vs II\end{tabular}} \\ \cmidrule(lr){3-5} \cmidrule(lr){6-8}
\multicolumn{1}{c}{}                          & \multicolumn{1}{c}{}                                                                      & \multicolumn{1}{c}{Avg.} & \multicolumn{1}{c}{Std. dev.} & \multicolumn{1}{c}{Best} & \multicolumn{1}{c}{Avg.} & \multicolumn{1}{c}{Std. dev.} & \multicolumn{1}{c}{Best} & \multicolumn{1}{c}{}                           &                                                                            \\ \midrule
R.200.100.1                                   & 61                                                                                        & 71.8                     & 2.8                           & 67                       & 63                       & 0.9                           & 62                       & \textless0.0001                                & II                                                           \\
R.200.100.15                                  & \textbf{1792}                                                                             & 1915.7                   & 33.6                          & 1849                     & 1821.6                   & 47.3                          & \textbf{1792}            & \textless0.0001                                & II                                                           \\
R.200.100.30                                  & \textbf{4216}                                                                             & 4216                     & 0                             & \textbf{4216}            & 4216                     & 0                             & \textbf{4216}            & 1                                              & -                                                                          \\
R.200.100.60                                  & \textbf{71749}                                                                            & 71749                    & 0                             & \textbf{71749}           & 71749                    & 0                             & \textbf{71749}           & 1                                              & -                                                                          \\
R.200.1000.1                                  & \textbf{1404}                                                                             & 1445.4                   & 15.3                          & 1408                     & 1423.2                   & 6.8                           & 1407                     & \textless0.0001                                & II                                                           \\
R.200.1000.15                                 & \textbf{20481}                                                                            & 21586.1                  & 373.6                         & 20837                    & 20639.2                  & 200.2                         & \textbf{20481}           & \textless0.0001                                & II                                                           \\
R.200.1000.30                                 & \textbf{41196}                                                                            & 41196                    & 0                             & \textbf{41196}           & 41196                    & 0                             & \textbf{41196}           & 1                                              & -                                                                          \\
R.200.1000.60                                 & \textbf{71556}                                                                            & 71556                    & 0                             & \textbf{71556}           & 71556                    & 0                             & \textbf{71556}           & 1                                              & -                                                                          \\ \midrule
R.300.100.1                                   & \textbf{26}                                                                               & 43.8                     & 5                             & 37                       & 27.8                     & 1.7                           & \textbf{26}              & \textless0.0001                                & II                                                           \\
R.300.100.15                                  & 3161                                                                                      & 3297.7                   & 29.3                          & 3238                     & 3174.2                   & 47.7                          & {\ul \textbf{3152}}      & \textless0.0001                                & II                                                           \\
R.300.100.30                                  & \textbf{6120}                                                                             & 6120.9                   & 1                             & \textbf{6120}            & 6120                     & 0                             & \textbf{6120}            & \textless0.0001                                & II                                                           \\
R.300.100.60                                  & \textbf{9726}                                                                             & 9726                     & 0                             & \textbf{9726}            & 9726                     & 0                             & \textbf{9726}            & 1                                              & -                                                                          \\
R.300.1000.1                                  & \textbf{1294}                                                                             & 1398.3                   & 26.9                          & 1356                     & 1394.1                   & 19.9                          & 1355                     & 0.5691                                         & -                                                                          \\
R.300.1000.15                                 & 29183                                                                                     & 31132.5                  & 542.8                         & 30013                    & 29309.7                  & 193.5                         & {\ul \textbf{29026}}     & \textless0.0001                                & II                                                           \\
R.300.1000.30                                 & \textbf{54147}                                                                            & 54179.3                  & 50.8                          & \textbf{54147}           & 54147                    & 0                             & \textbf{54147}           & \textless0.0001                                & II                                                           \\
R.300.1000.60                                 & \textbf{109471}                                                                           & 109490.8                 & 45.1                          & \textbf{109471}          & 109471                   & 0                             & \textbf{109471}          & 0.0214                                         & II                                                           \\ \midrule
R.400.100.1                                   & \textbf{13}                                                                               & 33.2                     & 4.2                           & 25                       & 14.6                     & 1.9                           & \textbf{13}              & \textless0.0001                                & II                                                           \\
R.400.100.15                                  & 3906                                                                                      & 4184.1                   & 50.4                          & 4084                     & 3902.6                   & 12.6                          & {\ul \textbf{3883}}      & \textless0.0001                                & II                                                           \\
R.400.100.30                                  & \textbf{8165}                                                                             & 8165.9                   & 0.3                           & \textbf{8165}            & 8165                     & 0                             & \textbf{8165}            & \textless0.0001                                & II                                                           \\
R.400.100.60                                  & \textbf{15228}                                                                            & 15228                    & 0                             & \textbf{15228}           & 15228                    & 0                             & \textbf{15228}           & 1                                              & -                                                                          \\
R.400.1000.1                                  & \textbf{1343}                                                                             & 1471.3                   & 24.7                          & 1426                     & 1497                     & 25                            & 1454                     & 0.0002                                         & I                                                                  \\
R.400.1000.15                                 & 43268                                                                                     & 41821.2                  & 504.7                         & 40869                    & 39207.1                  & 148                           & {\ul \textbf{38963}}     & \textless0.0001                                & II                                                           \\
R.400.1000.30                                 & \textbf{85128}                                                                            & 85253.2                  & 105.6                         & 85146                    & 85128                    & 0                             & \textbf{85128}           & \textless0.0001                                & II                                                           \\
R.400.1000.60                                 & \textbf{140816}                                                                           & 140816                   & 0                             & \textbf{140816}          & 140816                   & 0                             & \textbf{140816}          & 1                                              & -                                                                          \\ \midrule
R.500.100.1                                   & \textbf{4}                                                                                & 24                       & 4                             & 13                       & 4.9                      & 0.9                           & \textbf{4}               & \textless0.0001                                & II                                                           \\
R.500.100.15                                  & 5361                                                                                      & 5739                     & 76.8                          & 5578                     & 5348.7                   & 27                            & {\ul \textbf{5284}}      & \textless0.0001                                & II                                                           \\
R.500.100.30                                  & \textbf{9665}                                                                             & 9696.1                   & 9.1                           & 9687                     & 9667.6                   & 1                             & \textbf{9665}            & \textless0.0001                                & II                                                           \\
R.500.100.60                                  & \textbf{18240}                                                                            & 18255.9                  & 4.3                           & \textbf{18240}           & 18240                    & 0                             & \textbf{18240}           & \textless0.0001                                & II                                                           \\
R.500.1000.1                                  & \textbf{1316}                                                                             & 1465.6                   & 30.9                          & 1416                     & 1490.7                   & 31.5                          & 1438                     & 0.0043                                         & I                                                                  \\
R.500.1000.15                                 & 50725                                                                                     & 53818.4                  & 726.2                         & 52549                    & 49815.3                  & 197.8                         & {\ul \textbf{49504}}     & \textless0.0001                                & II                                                           \\
R.500.1000.30                                 & \textbf{98987}                                                                            & 99240.5                  & 117.9                         & 99018                    & 98987                    & 0                             & \textbf{98987}           & \textless0.0001                                & II                                                           \\
R.500.1000.60                                 & \textbf{178212}                                                                           & 178259.5                 & 123.1                         & \textbf{178212}          & 178212                   & 0                             & \textbf{178212}          & 0.0418                                         & II                                                           \\ \midrule
R.600.100.1                                   & \textbf{1}                                                                                & 17.4                     & 3                             & 12                       & 1.7                      & 1                             & \textbf{1}               & \textless0.0001                                & II                                                           \\
R.600.100.15                                  & 5684                                                                                      & 6143.7                   & 114.4                         & 5919                     & 5544                     & 30.1                          & {\ul \textbf{5493}}      & \textless0.0001                                & II                                                           \\
R.600.100.30                                  & \textbf{12465}                                                                            & 12484.6                  & 13.7                          & 12468                    & 12465                    & 0                             & \textbf{12465}           & \textless0.0001                                & II                                                           \\
R.600.100.60                                  & \textbf{23293}                                                                            & 23293                    & 0                             & \textbf{23293}           & 23293                    & 0                             & \textbf{23293}           & 1                                              & -                                                                          \\
R.600.1000.1                                  & \textbf{1337}                                                                             & 1515.7                   & 27.7                          & 1446                     & 1540.2                   & 25.6                          & 1492                     & 0.001                                          & I                                                                  \\
R.600.1000.15                                 & 57237                                                                                     & 60100.8                  & 785.7                         & 58426                    & 55891.4                  & 446.3                         & {\ul \textbf{55213}}     & \textless0.0001                                & II                                                           \\
R.600.1000.30                                 & \textbf{126798}                                                                           & 127106.1                 & 339.8                         & \textbf{126798}          & 126798                   & 0                             & \textbf{126798}          & \textless0.0001                                & II                                                           \\
R.600.1000.60                                 & \textbf{214608}                                                                           & 214608                   & 0                             & \textbf{214608}          & 214608                   & 0                             & \textbf{214608}          & 1                                              & -                                                                          \\ \midrule
R.700.100.1                                   & \textbf{1}                                                                                & 10.9                     & 2.2                           & 7                        & 1                        & 0                             & \textbf{1}               & \textless0.0001                                & II                                                           \\
R.700.100.15                                  & 7311                                                                                      & 7707                     & 90.4                          & 7560                     & 7125.3                   & 52.3                          & {\ul \textbf{7021}}      & \textless0.0001                                & II                                                           \\
R.700.100.30                                  & \textbf{14510}                                                                            & 14563.9                  & 24.7                          & 14516                    & 14510                    & 0                             & \textbf{14510}           & \textless0.0001                                & II                                                           \\
R.700.100.60                                  & \textbf{24102}                                                                            & 24103.7                  & 4.3                           & \textbf{24102}           & 24102                    & 0                             & \textbf{24102}           & 0.0215                                         & II                                                           \\
R.700.1000.1                                  & \textbf{1231}                                                                             & 1463                     & 34.6                          & 1382                     & 1484.9                   & 29.7                          & 1423                     & 0.028                                          & I                                                                  \\
R.700.1000.15                                 & 66837                                                                                     & 70778.6                  & 784.6                         & 69410                    & 65935.9                  & 436.9                         & {\ul \textbf{65305}}     & \textless0.0001                                & II                                                           \\
R.700.1000.30                                 & \textbf{134474}                                                                           & 135480                   & 339.9                         & 134942                   & 134474                   & 0                             & \textbf{134474}          & \textless0.0001                                & II                                                           \\
R.700.1000.60                                 & \textbf{245589}                                                                           & 245589                   & 0                             & \textbf{245589}          & 245589                   & 0                             & \textbf{245589}          & 1                                              & -                                                                          \\ \bottomrule
\end{tabular}%
}
\end{table}

\begin{table}[]
\centering
\caption{\R
Results of the EACS with the SOP-3-exchange (I) and EACS-SA with the SOP-3-exchange-SA (II) algorithms for the selected SOPLIB2006 instances with the time limit set to 1200 sec.
The meaning of the columns is as before.
}
\label{tab:cmp-ext-1200}
\resizebox{\textwidth}{!}{%
\begin{tabular}{@{}lrrrrrrrrc@{}}
\toprule
\multicolumn{1}{c}{\multirow{2}{*}{Instance}} & \multicolumn{1}{c}{\multirow{2}{*}{\begin{tabular}[c]{@{}c@{}}Best known\\\cite{a12,a13,papapanagiotou2015comparison}\end{tabular}}} & \multicolumn{3}{c}{EACS + SOP-3-exchange (I)} & \multicolumn{3}{c}{EACS-SA + SOP-3-exchange-SA (II)} & \multicolumn{1}{c}{\multirow{2}{*}{$p$-value}} & \multirow{2}{*}{\begin{tabular}[c]{@{}c@{}}Verdict\\ I vs II\end{tabular}} \\ \cmidrule(lr){3-5} \cmidrule(lr){6-8}
\multicolumn{1}{c}{} & \multicolumn{1}{c}{} & \multicolumn{1}{c}{Avg.} & \multicolumn{1}{c}{Std. dev.} & \multicolumn{1}{c}{Best} & \multicolumn{1}{c}{Avg.} & \multicolumn{1}{c}{Std. dev.} & \multicolumn{1}{c}{Best} & \multicolumn{1}{c}{} &  \\ \midrule
R.300.1000.1 & 1294 & 1397.7 & 27.9 & 1360 & 1361.3 & 13.7 & 1339 & \textless0.0001 & II \\
R.400.1000.1 & 1343 & 1456.0 & 26.1 & 1414 & 1462.1 & 24.9 & 1421 & 0.3254 & -- \\
R.500.1000.1 & 1316 & 1463.1 & 28.8 & 1416 & 1459.1 & 20.7 & 1411 & 0.7006 & -- \\
R.600.1000.1 & 1337 & 1504.2 & 30.2 & 1455 & 1508.8 & 26.0 & 1447 & 0.3477 & -- \\
R.700.1000.1 & 1231 & 1445.6 & 26.3 & 1398 & 1430.2 & 24.1 & 1381 & 0.0281 & II \\ \bottomrule
\end{tabular}%
}
\end{table}

\subsection{Speed comparison}
\label{sec:Speed_comparison}

\begin{figure}[!ht]
    \centering
    \includegraphics{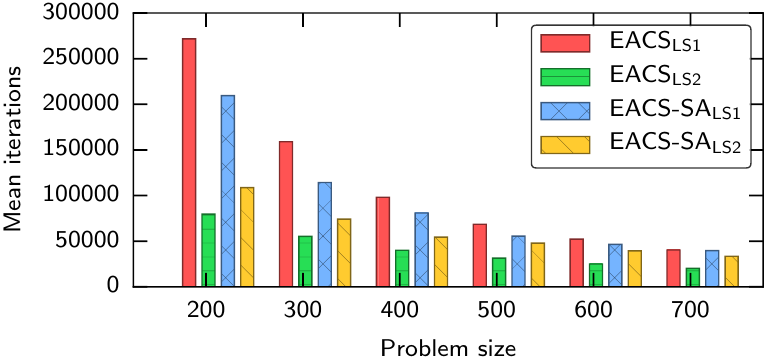}
    \caption{
    Average number of iterations performed by the 
    EACS and EACS-SA algorithms
    vs the size of the SOP instance.
    The \emph{LS1} and \emph{LS2} subscripts denote the local search method used, i.e.
    the \emph{SOP-3-exchange} and \emph{SOP-3-exchange-SA}, respectively.
    }
    \label{fig:eacs-cmp-iter}
\end{figure}

The algorithms differ not only in the quality of the generated solutions but also in the relative speed.
The SA component does not affect the asymptotic time complexity of the ACS and EACS but it may influence the solution search "trajectory", thus possibly impacting the runtime, particularly if a local search is used.
The SOP-3-exchange tries to improve a solution by searching only for the improving changes (moves) and its time complexity depends on the relative order of nodes in the solution. If the solution changes slightly from iteration to iteration, the runtime shortens because of the focusing only on the changed parts of the solution.
In contrast, the SOP-3-exchange-SA, due to the SA component, may also accept a number of worse (up-hill) moves, hence the overall runtime should increase.
Figure~\ref{fig:eacs-cmp-iter} shows a bar plot of the average number of iterations performed for the EACS and EACS-SA with both LS variants vs the size of the SOP instance.
As expected, the algorithms with the SOP-3-exchange-SA were slower than the algorithms with the SOP-3-exchange. The fastest algorithm was the EACS with the SOP-3-exchange, beating the EACS-SA with the same LS.
Interestingly, the slowest algorithm was the EACS with the SOP-3-exchange-SA; it was even slower than the EACS-SA with the same LS. This is probably due to the fact that the EACS can relatively easily get trapped in a "deep" local minimum from which an escape is difficult even if the SOP-3-exchange-SA accepts a number of up-hill moves.
On the other hand, the EACS-SA focuses on a larger number of different solutions during the search, some of which are less time-consuming to improve by the LS.
Finally, the larger the size of the instance, the lower the number of iterations performed by the algorithms.

\section{Summary}
\label{sec:Summary}

The Ant Colony System and particularly its enhanced version (EACS) are competitive metaheuristics whose efficiency has been shown in a number of cases~\cite{a9,a11,a12}.
Nevertheless, we have shown that the search process of the ACS and EACS can still be improved with ideas taken from Simulated Annealing. Specifically, instead of increasing the pheromone values based on the current best solution, the proposed \emph{ACS-SA} and \emph{EACS-SA} algorithms increase the pheromone values based on the current \emph{active solution} that is chosen from among all the solutions constructed by the ants. The active solution may not necessarily be the current best solution as it is selected probabilistically by using the Metropolis criterion from the SA.
This change weakens the exploitative focus of the ACS and EACS, thus increasing the chance of escaping local optima.  
The computational experiments on a set of SOP instances from the TSPLIB repository and subsequent statistical analyses have shown that in most cases the resulting ACS-SA and EACS-SA algorithms perform significantly better than the original algorithm.

An efficient local search heuristic is necessary for state-of-the-art performance in solving the SOP. 
Based on the same SA inspirations, we proposed an enhanced version of the state-of-the-art SOP-3-exchange heuristic by Gambardella~\cite{a11}. The resulting \emph{SOP-3-exchange-SA} algorithm is more resilient to getting trapped in local minima, at the expense of increased computation time.
The computational experiments conducted on a set of 48 SOP instances sized from 200 to 700 showed that the proposed EACS and EACS-SA with the SOP-3-exchange and SOP-3-exchange-SA local searches are in many cases able to find solutions of better quality than the original EACS with the SOP-3-exchange (a current state-of-the-art metaheuristic for the SOP), within the same computation time limit.
In fact, new, best solutions were obtained for 10 instances. In total, the best known or improved solutions were obtained at least once for a total of 85\% of the SOP instances considered here.

Although the proposed modifications are easy to implement and improve the performance of the original algorithms, they have some minor drawbacks. First, they increase the computation time relative to the original algorithms. Second, they require to set the values of the new parameters related to the SA cooling schedule ($\lambda$ and $\gamma$).
Also, relatively poor performance for SOP instances with a small number (1\%) of precedence constraints shows that there is still room for improvement, both in the ACS-SA, EACS-SA and local search methods.

In the future, a more advanced cooling schedule could be used to improve the convergence of the SA component of the proposed algorithms. 
A good candidate seems to be the adaptive cooling schedule that was proposed by Lam~\cite{a30}, although it requires a complex parameter setting and a method of controlling how much the subsequent solutions differ from one another.
An interesting idea could also be to activate the SA component only if search process stagnation is detected. 
Because the proposed fusion between the ACS and SA is problem-agnostic one could try to apply it to solve other difficult combinatorial optimization problems.
The performance of the proposed algorithms in terms of computation time could also be improved with the help of parallel computations, as the ACS is susceptible to parallelization even with modern GPUs~\cite{skinderowicz2016}.

\noindent
\textbf{Acknowledgments}: This research was supported in part by PL-Grid Infrastructure.


\bibliographystyle{plain}
\footnotesize
\bibliography{article}

\end{document}